\documentclass[11pt]{article}

\usepackage[final]{acl}

 \usepackage{microtype}

\usepackage[T1]{fontenc}    
\usepackage[utf8]{inputenc} 

\usepackage{newtxtext}      

\usepackage[english]{babel}

\definecolor{cvprblue}{rgb}{0.21,0.49,0.74}
\definecolor{greenx}{RGB}{0,128,128}
\definecolor{maroonx}{RGB}{195,18,48}

\usepackage{hyperref} 
\hypersetup{
    colorlinks=true,
    linkcolor=maroonx,
    anchorcolor=blue,
    citecolor=cvprblue,
    pagebackref=true
}

\usepackage{microtype}
\usepackage{xurl}
\usepackage{booktabs}
\usepackage{enumitem}
\usepackage{multicol}
\usepackage{CJKutf8}
\usepackage{amsmath}
\usepackage{siunitx}
\usepackage{floatflt}
\usepackage{graphicx}
\usepackage{booktabs}
\usepackage{wrapfig}
\usepackage{lipsum}
\usepackage{algorithm}
\usepackage{algorithmicx}
\usepackage{algpseudocode}
\usepackage{microtype}
\usepackage{subfigure}
\usepackage{multirow}
\usepackage{pifont}  
\usepackage{amssymb} 
\usepackage{makecell}
\usepackage[colorinlistoftodos]{todonotes}
\usepackage[table]{xcolor}
\usepackage[most]{tcolorbox}
\usepackage{lipsum} 
\usepackage[most]{tcolorbox}
\usepackage[most]{tcolorbox} 
\usepackage{listings}        
\usepackage{xcolor}          
\usepackage{lipsum}          
\usepackage{xcolor}
\usepackage{booktabs}   
\usepackage{multirow}   
\usepackage{caption}    

\newtcolorbox{userbox}{
    colback=blue!5!white, colframe=blue!75!black, fonttitle=\bfseries,
    colbacktitle=blue!20!white, coltitle=black, title=User, arc=2mm, boxrule=1pt,
}
\newtcolorbox{agentbox}{
    colback=black!5!white, colframe=black!75!black, fonttitle=\bfseries,
    colbacktitle=black!20!white, coltitle=black, title=Agent, arc=2mm, boxrule=1pt,
}

\definecolor{codegreen}{rgb}{0,0.6,0}
\definecolor{codegray}{rgb}{0.5,0.5,0.5}
\definecolor{codepurple}{rgb}{0.58,0,0.82}
\definecolor{backcolour}{rgb}{0.95,0.95,0.95}

\lstdefinestyle{mystyle}{
    backgroundcolor=\color{backcolour},   
    commentstyle=\color{codegreen},
    keywordstyle=\color{magenta},
    numberstyle=\tiny\color{codegray},
    stringstyle=\color{codepurple},
    basicstyle=\ttfamily\footnotesize,
    breakatwhitespace=false,         
    breaklines=true,                 
    captionpos=b,                    
    keepspaces=true,                 
    numbers=left,                    
    numbersep=5pt,                  
    showspaces=false,                
    showstringspaces=false,
    showtabs=false,                  
    tabsize=2
}
\newtcolorbox{codeblock}[2][]{
    listing engine=listings,
    listing options={
        language=#1,
        style=mystyle 
    },
    colback=green!10!white,
    colframe=black!60!black,
    boxrule=0.5pt,
    arc=1mm,
    title=#2,
    fonttitle=\bfseries\ttfamily\small,
    colbacktitle=black!20!white,
    coltitle=black,
    left=3mm, right=3mm, top=1mm, bottom=1mm,
}

\newtcolorbox{expblock}[2][]{
    listing engine=listings,
    listing options={
        language=#1,
        style=mystyle 
    },
    colback=orange!10!white,
    colframe=black!60!black,
    boxrule=0.5pt,
    arc=1mm,
    title=#2,
    fonttitle=\bfseries\ttfamily\small,
    colbacktitle=black!20!white,
    coltitle=black,
    left=3mm, right=3mm, top=1mm, bottom=1mm,
}

\usepackage{array}
\usepackage{amsmath}
\usepackage{amssymb}
\usepackage{mathtools}
\usepackage{amsthm}
\newlist{todolist}{itemize}{2}
\setlist[todolist]{label=$\square$}
\usepackage[capitalize,noabbrev]{cleveref}
\usepackage{adjustbox} 

\usepackage[utf8]{inputenc}
\usepackage{caption} 
\usepackage{adjustbox} 
\usepackage{arydshln}
\usepackage{fontawesome}
\usepackage{subcaption}
\usepackage{tcolorbox}
\tcbuselibrary{skins,breakable}
\newcommand{\yes}{\ding{51}}
\newcommand{\no}{\ding{55}}
\tcbuselibrary{skins}

\newtcolorbox{chatconversation}[2][]{
    breakable, 
    colback=white,
    colframe=black!75!white,
    boxrule=1pt,
    arc=3mm,
    title=#2, 
    label=#1, 
    fonttitle=\bfseries,
    colbacktitle=black!10!white,
    coltitle=black,
    attach boxed title to top center={yshift=-3mm},
    boxed title style={
        colback=white,
        colframe=white,
    },
    underlay={
        \node[anchor=north, font=\bfseries] at (title.south) {};
    }
}

\NewDocumentCommand{\hongru}
{ mO{} }{\textcolor{blue}{\textsuperscript{\textit{Hongru}}\textsf{\textbf{\small[#1]}}}}

\title{WebAggregator: Enhancing Compositional Reasoning Capabilities of \\Deep Research Agent Foundation Models}

\author{
Rui Wang\textsuperscript{\dag}, Ce Zhang, Jun-Yu Ma, Jianshu Zhang, Hongru Wang\textsuperscript{\dag}, Yi Chen, Boyang Xue\textsuperscript{\dag}, \\
\textbf{Tianqing Fang\textsuperscript{\ddag*}, Zhisong Zhang\textsuperscript{\ddag}, Hongming Zhang\textsuperscript{\ddag}, Haitao Mi\textsuperscript{\ddag}, Dong Yu\textsuperscript{\ddag}, Kam-Fai Wong\textsuperscript{\dag\thanks{Correspondence to Rui 
~\url{ruiwangnlp@outlook.com}, Tianqing~\url{fangtq229@gmail.com}, Kam-Fai~\url{kfwong@cuhk.edu.hk} }}} \\
\textsuperscript{\dag}MoE Lab, The Chinese University of Hong Kong 
\textsuperscript{\ddag}Tencent AI Lab\\
\faGithub ~\url{https://github.com/Tencent/CognitiveKernel-Pro}  \\
\faGithub ~\url{https://github.com/Tencent/WebAggregator}  \\
}

\begin{document}

\maketitle
\begin{abstract}

The hallmark of Deep Research agents lies in compositional reasoning, the capacity to aggregate distributed, heterogeneous information into coherent logical insights. 
However, current agentic systems are often retrieval-heavy but reasoning-light, where success is predominantly determined by simple entity-seeking rather than the multi-step aggregation of scattered evidence. 
To address this, we propose a data synthesis pipeline \textbf{WebAggregator}, designed to shift the agentic paradigm from retrieval-centric to compositional aggregation.                                           
Our approach first employs \textit{Proactive Explorer} to collect interconnected knowledge, then \textit{Compositional Logic Proposer} to weave knowledge into complex questions using over 12 composition guidelines derived from a rigorous deconstruction of the Deep Research problem setting.
By leveraging 10K verifiable QA pairs grounded on 50K websites, we curate a high-quality SFT dataset via rejection sampling. 
Fine-tuning on this corpus fundamentally transforms agent behavior, fostering deliberate composition reasoning and reduced tool redundancy.
The resulting WebAggregator-32B surpasses GPT-4.1 and matches Claude-3.7-Sonnet on GAIA, WebWalkerQA, and XBench. 
To address the lack of benchmarks that emphasize both reasoning and retrieval, we introduce the WebAggregatorQA testbed, which reveals that even with perfect retrieval, top-tier models still underperformed. These results demonstrate that compositional reasoning, not retrieval, is the true performance ceiling for next-generation research agents.

\end{abstract}

\section{Introduction}

Deep Research agent systems~\citep{deepresearch,manus} aim to automate scientific discovery using foundational large language models (LLMs). 
Achieving this requires not merely effective \textit{information seeking}, grounding responses with accurate knowledge, but more critically, \textit{compositional reasoning}: the ability to integrate distributed materials into coherent insights by decoding underlying logical structures~\citep{psychology,cog-of-write, gaia}.

\begin{figure}
    \centering
    \includegraphics[width=1.0\linewidth]{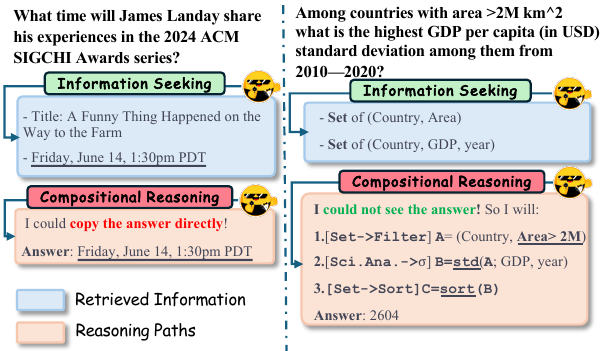}
    \caption{Comparison between WebWalkerQA (\textbf{left})~\cite{wu2025webwalkerbenchmarkingllmsweb} and WebAggregatorQA (\textbf{right, ours}).
    WebAggregatorQA necessitates both \textit{Information Seeking} (IS) and intensive \textit{Compositional Reasoning} load, which reveals a critical finding: \textbf{\textit{accurate retrieval could not guarantee task success}}. 
    }
    \label{fig:teaser}
\end{figure}

\begin{figure*}[htbp] 
\centering 


{\small 
\begin{tabular}{l|c|c|c|c|c}
\hline
\textbf{Resource} & \textbf{IS} & \textbf{CR} & \textbf{Train set} & \textbf{Information Source} &\textbf{\#Dom.} \\
\hline
{GAIA~(\citeauthor{gaia}}) & {\yes} & {\yes} & {N/A} & Text, Figure, Audio, File, Dynamic Elements & - \\
BrowseComp~(\citeauthor{browsercomp-en}) & {\yes}  & \no & N/A & Text &9\\
WebWalker~(\citeauthor{wu2025webwalkerbenchmarkingllmsweb}) & \yes & \no & Y & Text &4 \\
TaskCraft~(\citeauthor{shi_taskcraft_2025}) & \yes & \no & Y & Text, PDF, Figure &8\\
WebShaper~(\citeauthor{tao2025webshaper}) & \yes & \no & N & Text & 11\\
\hdashline
\textbf{{WebAggregatorQA (Ours)}} & {\yes} & {\yes} & {Y} & Text, Figure, File, Dynamic Elements & {12}  \\ 
\hline
\end{tabular}
}

\vspace{0.4em} 

\includegraphics[width=1.0\linewidth]{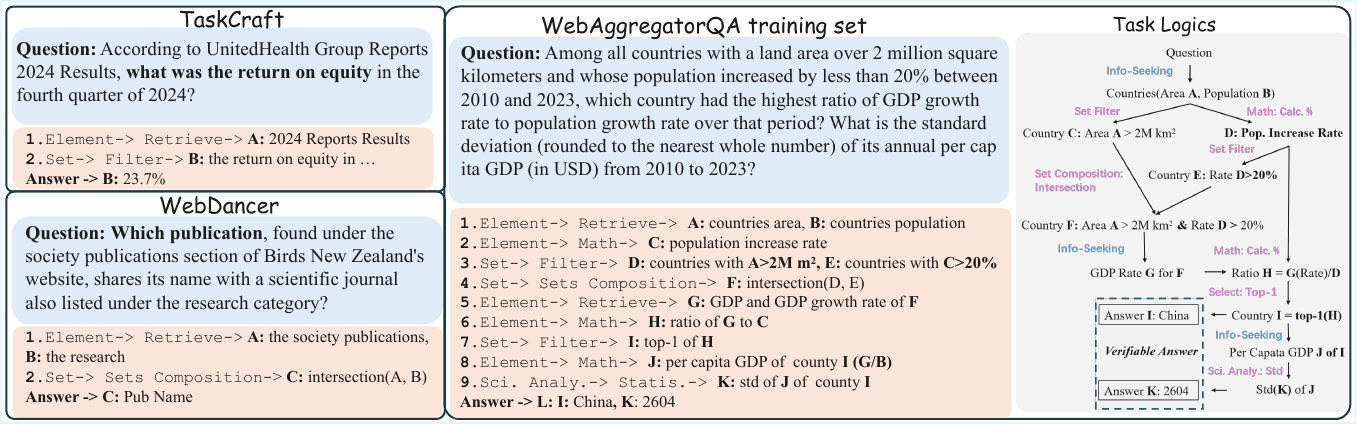}


\captionof{figure}{
    \textbf{(Top)} The table illustrates a feature comparison between our benchmark and existing resources. It highlights our dataset's unique focus on \textcolor[HTML]{DC70CF}{\textbf{Compositional Reasoning (CR)}} as a progression beyond traditional \textcolor[HTML]{66A2DC}{\textbf{Information-Seeking (IS)}} needs.
    \textbf{(Bottom)} Examples comparison of the simple logic in prior works with the complex compositional reasoning demanded by WebAggregatorQA.}
\label{tab:resource-info} 
\vspace{-4mm}
\end{figure*}

However, current resources fall short of fostering this capability. 
Traditional multi-hop datasets~\citep{yang-etal-2018-hotpotqa} offer only shallow logical hops, while recent web agent benchmarks~\citep{shi_taskcraft_2025,wu2025webwalkerbenchmarkingllmsweb} tend to prioritize entity-seeking puzzles on static pages. 
There are two fundamental gaps: (1) insufficient environmental complexity, such as the dynamic content (e.g., JavaScript) and files essential for research; and (2) an imbalance in task difficulty, which relies on search effort rather than reasoning depth, as shown in Figure~\ref{fig:teaser} \& \ref{tab:resource-info}. 

Specifically, $30.79\%$ of WebWalkerQA~\citep{deng2023mind2web} and $43.2\%$ of TaskCraft~\citep{shi_taskcraft_2025} tasks only require simple entity localization. 
This observation aligns with other resources~\citep{wu_webdancer_2025,tao2025webshaper}, suggesting that existing tasks may not fully test high-level reasoning.
This tendency encourages agents to rely on frequent tool execution rather than necessary reasoning steps (Section~\ref{sec:core-behavior}), leaving compositional reasoning as a critical but underexplored frontier.

\begin{figure*}
    \centering
    \includegraphics[width=1.0\linewidth]{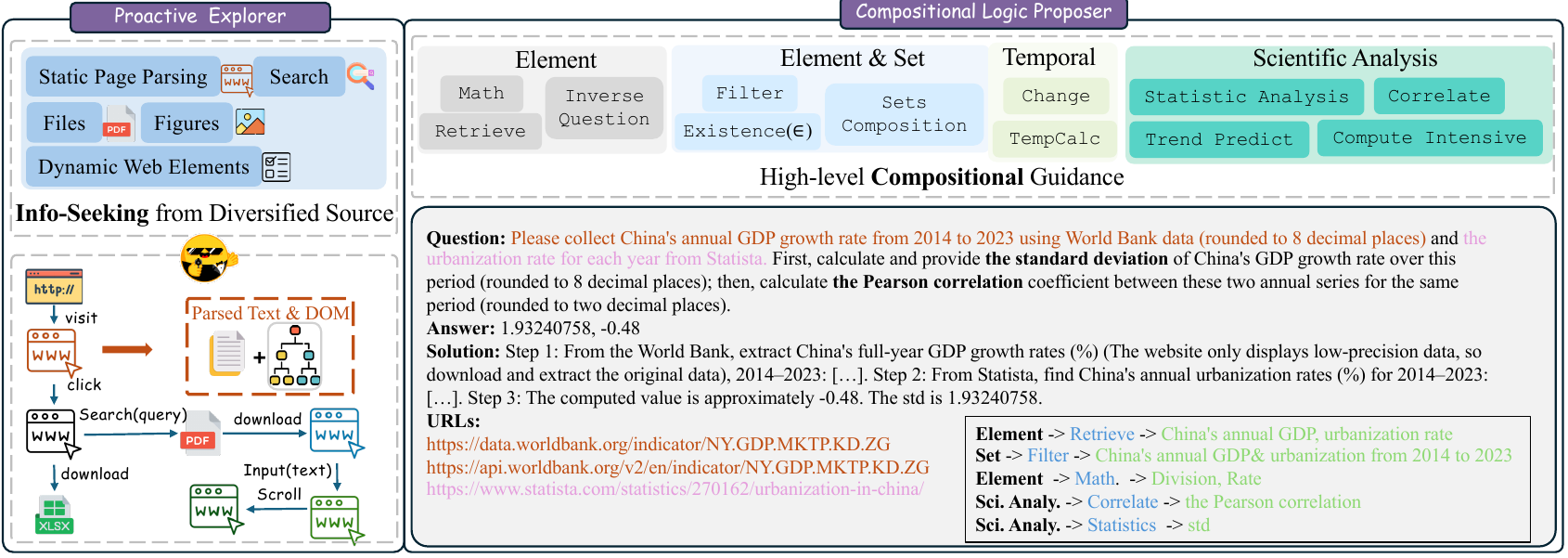}
    \caption{\textbf{Pipeline of WebAggregator for WebAggregatorQA construction.} {(1) Proactive Explorer} gathers comprehensive information by interacting with the web environment through tools (more details in Figure~\ref{fig:web-explore}). {(2) Compositional Logic Proposer} synthesizes QA pairs based on the collected knowledge by instantiating the high-level composition guidance into concrete operations, e.g., \textit{Statistic Analysis} $\rightarrow$ \textit{standard deviation}.     }
    \label{fig:basic-illus}
    \vspace{-3mm}
\end{figure*}

To bridge these gaps, we introduce \textbf{WebAggregator}, an automated framework designed to generate verifiable training data that pushes agents beyond information seeking towards deep {co}mpositional {re}asoning (Figure~\ref{fig:basic-illus}) . 
The workflow consists of two phases: the \textit{Proactive Explorer} and the \textit{Compositional Logic Proposer}.
Equipped with a comprehensive toolkit, spanning search, static parsing, dynamic interaction, file processing, and visual perception, the agent is capable of navigating in realistic web environments.
\textit{Proactive Explorer} initiates exploration from anchor URLs to autonomously gather interconnected resources.
Subsequently, the \textit{Compositional Logic Proposer} synthesizes complex tasks that explicitly necessitate multi-hop aggregation operations across these collected materials.
Drawing on prior multi-hop reasoning studies~\citep{DBLP:conf/cvpr/ChangCNGSB22WebQA, yang-etal-2018-hotpotqa,talmor_multimodalqa_2021,wu2025tablebench,widesearch}, this module employs a rigorous taxonomy categorized into Element, Set, Scientific Analysis, and Temporal Reasoning, which are further refined into 12 granular subtypes (Table~\ref{tab:categories}) to tailor the system to the analytical depth required by Deep Research.

We utilize WebAggregator to construct WebAggregatorQA, a dataset of approximately 10K QA pairs designed to enforce rigorous compositional reasoning. 
As illustrated in Figure~\ref{tab:resource-info}, WebAggregatorQA are designed to be non-trivial. Even with all necessary references provided, an agent must perform genuine compositional reasoning to derive the answer, rather than relying on direct extraction. 
Our rigorous quality control ensures the resulting tasks demand genuine compositional reasoning rather than shallow pattern matching.
By leveraging rejection sampling to train the WebAggregator model family, we demonstrate the high effectiveness of this training corpus. 
Extensive experiments on GAIA, WebWalkerQA, and XBench show that WebAggregator-8B significantly surpasses GPT-4.1, while the 32B variant outperforms all strong baselines. 
Furthermore, fine-tuning on WebAggregatorQA fundamentally improves agent behavior by encouraging deeper reasoning steps and clearly reducing redundant tool usage, leading to robust performance gains.

The manually annotated WebAggregatorQA test set serves as a challenging benchmark, where even Claude-3.7-Sonnet achieves only 28.3\%. 
Crucially, our error analysis reveals a critical limitation: \textit{accurate reference retrieval does not guarantee task success}. 
Even with the full reference context provided, Claude-3.7-Sonnet scores only 42.1\%, highlighting severe challenges in compositional reasoning.
Our further analysis attributes this failure to gaps in advanced agentic capabilities, such as ineffective self-refinement, and faulty information utilization.
These findings underscore the enduring value of WebAggregatorQA as a testbed, suggesting that future progress must transcend retrieval-centric paradigms to foster agents capable of the rigorous compositional reasoning required for complex tasks.

\section{WebAggregator}

We aim to automatically synthesize scalable, challenging QA pairs grounded in real web resources, requiring \textit{minimal human involvement}. 
Unlike fact-lookup puzzles~\cite{browsercomp-en,wu2025webwalkerbenchmarkingllmsweb}, these tasks demand both complex retrieval and compositional reasoning. 
We propose an automatically verifiable pipeline, WebAggregator (Figure~\ref{fig:basic-illus}), that frames synthesis as an agentic task:\footnote{Agent details in Appendix~\ref{app:agent-st}.} starting from anchor URLs, \textbf{Proactive Explorer} harvests interconnected information from diversified sources, which is formulated by \textbf{Compositional Logic Proposer} into reasoning-intensive QAs, which then undergo rigorous automated quality control.

\subsection{Anchor URL Collection}

By selecting seed URLs across multiple domains, we can effectively regulate the dataset’s domain distribution, thereby enabling effective distribution balance.
We sampled 5,000 topic-diverse queries from QA and scientific datasets~\citep{yang-etal-2018-hotpotqa,pubmedqa,trivedi2021musique} and retrieved URLs via Google Search, resulting in over 160,000 anchor links from 11+ domains (Figure~\ref{fig:dist-domain-step}).

\subsection{WebAggregator: From Exploration to Composition}

In this section, we introduce our approach to synthesizing target QA pairs by framing \textit{task synthesis as a specialized form of web-agent tasks}. The agent is given a task prompt consisting of two components: Proactive Explorer and Compositional Logic Proposal, enabling it to complete the task construction in an end-to-end manner.

\subsubsection{Proactive Explorer}
\textbf{Proactive Explorer} aims to collect diverse information as the foundation for task construction by exploration on the real web. 
During this phase (Figure~\ref{fig:basic-illus}), the agent is prompted to start from a single anchor URL and employ various tools to navigate across web pages to discover any unknown but relevant information. 
Interactions include navigating heterogeneous content types such as text, files, and images, as well as dynamic element interactions, which are still challenging for even state-of-the-art vision language models~\citep{VLM2}. 
To control task difficulty and ensure the comprehensiveness of the knowledge scope, a minimum number of web page visits is enforced (Appendix~\ref{app:data-cons-prompt}).

\begin{figure*}
    \centering
    \includegraphics[width=1\linewidth]{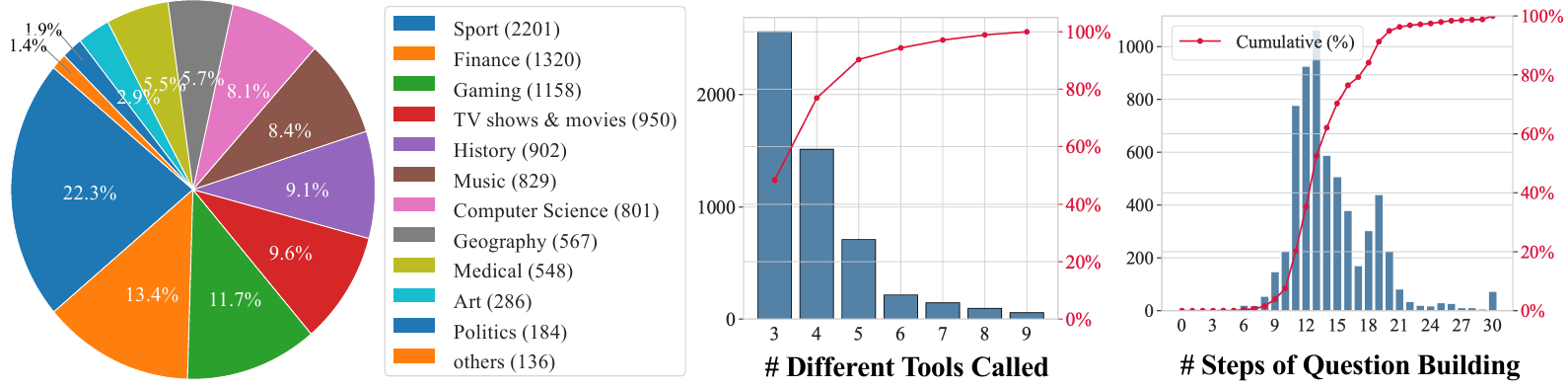}
    \caption{The distribution of domains in WebAggregatorQA, tasks categorized by the number of different tools involved during construction, and steps (an action-observation round) in data synthesis.}
    \label{fig:dist-domain-step}
    \vspace{-3mm}
\end{figure*}

This unconstrained exploration proves highly effective at sourcing diverse information. 
Analysis of 5,296 exploration trajectories (Figure~\ref{fig:dist-domain-step}) reveals significant interaction diversity. 
Every trajectory leverages at least three foundational tools: Search, Visit, and ScreenShot. 
Crucially, a majority of trajectories employ a wider toolkit: 48.36\% use 4 distinct tools, and 28.55\% use 5, while 13.41\% engage with 6 or more. 
This breadth of interaction enriches the knowledge pool and directly translates into more complex downstream tasks. 
For instance, questions derived from file-based information inherently test an agent's file-processing capabilities.

\subsubsection{Compositional Logic Proposer}

\textbf{Compositional Logic Proposer} transforms raw, unstructured information into sophisticated reasoning challenges. 
To systematically generate diverse and challenging reasoning chains, we developed a taxonomy of logical operations inspired by prior work on multi-hop and compositional QA~\citep{sen-etal-2022-mintaka,talmor_multimodalqa_2021,wu2025tablebench,DBLP:conf/acl/FangCSB24com2,frames}. 
As shown in Figure~\ref{fig:basic-illus}, this taxonomy is organized into four high-level categories: Element, Set, Scientific Analysis, and Temporal Reasoning, which are further refined into 12 granular subtypes.
Element and Set operations are the basis of compositional reasoning behaviors among knowledge snippets, such as \textit{mathematical calculation} among elements and \textit{Filtering} among sets.
While Scientific Analysis and Temporal Reasoning are advanced applications of them, involving computation-intensive and data science tasks, to reflect real but challenging user cases.

Crucially, these 12 subtypes act as high-level conceptual guides within the prompt, not as rigid templates.
They instruct the Proposer to instantiate these patterns into concrete, context-specific questions. For example, \textit{Element->mathematical calculation} guide can be instantiated as addition, subtraction, or ratio calculation, depending on the available data.
This flexible, guided approach ensures the generation of novel and diverse reasoning challenges (examples provided in Table~\ref{tab:categories}).

\subsubsection{Quality Control}

\noindent \textbf{QA Alignment Checking} We implement a two-stage verification process to ensure data integrity. First, the Proposer employs a self-refinement tool with a checklist to revise questions (Appendix~\ref{app:question-refine-tool}). Second, another agent reviews the data by solving it, verifying reference URLs to guarantee strict alignment between questions, answers, and sources (Appendix~\ref{app:prompt-refine-agent}). This rigorous post-hoc filtering removes 11.72\% of the initial samples.

\noindent \textbf{Diversity Constraint} We actively balance the dataset across domains and compositional types.\footnote{See Appendix~\ref{app:data-contain} for contamination analysis.} 
First, we normalize the distribution of anchor URL domains annotated by GPT-4.1 (Figure~\ref{fig:dist-domain-step}). Second, we dynamically adjust prompts to upweight underrepresented composition operations (e.g., calculating average) identified via preliminary analysis. 
As shown in Figure~\ref{fig:wc}, this strategy successfully spawns diverse low-level operations tailored to high-level guidance, such as intersection for Set tasks and table processing for Scientific Analysis.

\subsubsection{Trajectory Sampling}
After the task synthesis, we collect the trajectory that completes these tasks. 
We utilize the agent based on GPT-4.1 with SmolAgents and collect the plain text trajectories.
To ensure the quality of the collected trajectories, we conduct a further rejection sampling and finally collected 6,184 trajectories for the foundation model training.
(1) \textbf{\textit{Correctness}}: We retain those trajectories with correct predictions according to the reference answers. 
(2) \textbf{\textit{Format}}: Data with format errors (e.g., undefined tool name or parameters) is filtered out.
(3) \textbf{\textit{Exception Handling}}: Anomalies in observations (e.g., page failures) are kept to improve the model’s generalization, since similar situations would occur in the real web.

\begin{table*}[!t]
\centering
\small
\begin{tabular}{lcccc!{\vrule}cccc}
\toprule
\midrule

 \multirow{2}*{\textbf{Methods}} & \multicolumn{4}{c}{\textbf{GAIA-text}} & \multicolumn{4}{c}{\textbf{WebAggregatorQA}}\\

& \textbf{Level-1} & \textbf{Level-2} & \textbf{Level-3} & \textbf{Average}& \textbf{Level-1} & \textbf{Level-2} & \textbf{Level-3}  & \textbf{Average}\\
\midrule
\multicolumn{9}{l}{\cellcolor[RGB]{140, 210 255}{\emph{Non-Agentic}} }\\
 GPT-4.1 &10.3 &13.5 &8.3 &11.7  &15.4 &4.0 &2.8 & 5.6 \\
Claude-3.7-sonnet  &35.9 &17.3 &0.0 &22.3 &18.5 &5.1 &2.8  & 6.8\\
\hdashline

Qwen2.5-7B  & 12.8 & 3.8 & 0.0 & 6.8 & 4.2 & 1.0 & 0.0 &1.3 \\
Qwen2.5-32B  & 20.5 & 9.6 & 8.3 & 13.6& 4.2 & 1.0 & 0.0 &1.3 \\
\hdashline

Qwen3-8B & 12.8&3.8 &0.0 &6.8 & 4.2 & 1.0 & 2.8 &1.9\\
Qwen3-32B & 17.9&3.8 &0.0 &8.7 & 8.3 & 1.0 & 0.0 &1.9 \\
\hline

\multicolumn{9}{l}{\cellcolor[RGB]{140, 210 255}{\emph{Zero-shot Foundations}} }\\

 GPT-4.1  & 51.3 & 44.2 & 16.7 & 43.7 &62.4 &22.2  &{11.1} &25.8 \\
 {GPT-5} &{71.8} &{57.7} &{33.3} &{60.2} &{58.3} &{24.5} &\textbf{{22.2}} & \textbf{28.9} \\
Claude-3.7-sonnet  &\textbf{74.4} &\textbf{55.8} &\textbf{33.3} &\textbf{60.2} &\textbf{66.7} &\textbf{25.3}  &{11.1} & {28.3}\\
\hdashline

Qwen2.5-7B & 23.1 & 15.4 & 0.0 & 16.5 &27.3 &3.4 & 2.8&  6.3\\
Qwen2.5-32B  &46.1 & 21.2 & 0.0 & 28.2 &25.0 & 10.1 & 5.6 & 11.3  \\
\hdashline

Qwen3-8B & 33.3 & 11.5 & 0.0 & 18.4 & 30.8& 5.1 &5.6 & 9.4\\
Qwen3-32B & 48.7 & 40.4 & 16.7 & 40.8 & 45.8 & 10.1 & 5.6 & 14.5 \\
\hline
\multicolumn{9}{l}{\cellcolor[RGB]{140, 210 255}{\emph{Fine-tuned Foundations}} }\\

\multicolumn{9}{l}{\textit{\textbf{WebThinker}}} \\

Qwen2.5-32B & 56.4 & 50.0 & 16.7 &48.5 & \multicolumn{4}{c}{---}\\

\multicolumn{9}{l}{\textit{\textbf{WebDancer}}} \\

Qwen2.5-7B  &41.0 &30.7 & 0.0 & 31.0& \multicolumn{4}{c}{---} \\
Qwen2.5-32B  &46.1 &44.2 &8.3 & 40.7& \multicolumn{4}{c}{---} \\

\multicolumn{9}{l}{\textit{\textbf{WebSailor}}} \\

Qwen2.5-7B  &- &- & - & 37.9& \multicolumn{4}{c}{---} \\
Qwen2.5-32B  &- &- &- & 53.2& \multicolumn{4}{c}{---} \\

\multicolumn{9}{l}{\textit{\textbf{WebShaper}}} \\

Qwen2.5-32B  & 61.5 & 53.8 & 16.7 & 52.2& \multicolumn{4}{c}{---} \\

\multicolumn{9}{l}{\textit{\textbf{CogKernal-Pro}}} \\

Qwen3-8B  & 56.4 & 42.3 & 8.3 & 43.7& \multicolumn{4}{c}{---} \\
\hline

\multicolumn{9}{l}{\textit{\textbf{WebAggregator}}} \\
\rowcolor{gray!10}
Qwen2.5-7B & 53.8 & 30.8 & 16.7 &40.8 &37.5 & 11.1 & 8.3 & 14.5\\

\quad - \textit{pass@3} & {74.4} & {63.5} &25.0& {63.1} &54.2 &22.2 & 19.4 &26.4 \\

\rowcolor{gray!10}
Qwen2.5-32B & 66.7 & 44.2 & \textbf{33.3} & 51.5 &54.2 &15.2 & \textbf{11.1} & 20.1\\

\quad - \textit{pass@3} & {79.5} & {67.3} &50.0& {69.9} &70.8 &22.2 & 19.4 &28.9 \\

\hdashline
\rowcolor{gray!10}
Qwen3-8B  & 61.5 & 34.6 & 16.7 & 42.7 & 54.2 & {11.1} & 5.6 & 16.4 \\

\quad - \textit{pass@3} & {82.1} & {53.8} &33.3& {62.1} &62.4 &21.2 & 11.1 &25.2 \\
\rowcolor{gray!10}
Qwen3-32B & \textbf{69.2} & \textbf{55.8} &16.7& \textbf{56.3} &\textbf{62.4} &\textbf{24.2} & 8.3 &\textbf{26.4} \\
\quad - \textit{pass@3} & {79.5} & {67.3} &50.0& {69.9} &66.7 &35.4 & 13.9 &35.2 \\

\midrule
\bottomrule
\end{tabular}
\caption{The Pass@1 \& {Pass@3} performance of agents on GAIA-text and WebAggregatorQA. The best performance of different settings is in bold.}
\label{tab:related}
\vspace{-3mm}
\end{table*}

\subsubsection{Statistics of WebAggregatorQA}
WebAggregatorQA comprises 9,883 tasks (with 200 reserved for testing), covering 54,064 unique URLs across 12 domains. Figure~\ref{fig:dist-domain-step} shows the distribution of domains and steps for QA construction trajectories. 
Most QA pairs are constructed with around 15 steps, demonstrating that the generated data points are not hastily created from only a few reasoning steps, thus avoiding overly simplistic questions.

\subsection{WebAggregatorQA testset for Compositional Reasoning Evaluation}

Existing benchmarks~\citep{wu2025webwalkerbenchmarkingllmsweb,browsercomp-en} primarily assess information-seeking, often neglecting compositional reasoning. 
To address this critical gap, we introduce the test set of WebAggregatorQA, which is meticulously annotated to evaluate both retrieval and reasoning capabilities.

\noindent \textbf{Annotation Details} \quad
We uniformly split 200 tasks as seeds from WebAggregatorQA across different domains to ensure high task diversity. 
We aim to ensure every question is unambiguous, with the reference answer correct, other candidates are verifiably incorrect.
First, human annotators solve the QA pairs, then review the QA and references to eliminate ambiguities by adding clear constraints. 
To further enhance sample reliability and reduce bias from the solver's perspective, this process is repeated twice: tasks are solved, ambiguities identified, and revisions made by annotators.
In the final cross-validation stage, each question was answered by at least two annotators, yielding 155 consistently aligned samples. 
The other 4 samples were abandoned by annotators due to difficulty, but their references and questions were verified to ensure data quality and thus were retained.
More details are shown in Appendix~\ref{app:test-anno}.

This yielded 159 samples, including those in text and multimodal, categorized by difficulty into Level 1 (24), Level 2 (99), and Level 3 (36). 
Each sample contains a question, reference answer, solution, and supporting URLs.
A text example is shown in Figure~\ref{tab:resource-info} and a multimodal one is in Figure~\ref{fig:gaia-web-illus-mm}.

\section{Experiments}

\subsection{Experimental Setups} 
\noindent \textbf{Models and Benchmarks} \quad
We construct the WebAggregator models by SFT varied Qwen2.5~\citep{qwen2.5} and Qwen3~\citep{qwen3} models on the training set of WebAggregatorQA.
We evaluate the baselines and our methods on the GAIA-text~\citep{gaia} following~\citet{li_websailor_2025,wenthinker}, XBench~\cite{xbench},  WebWalkerQA~\cite{wu2025webwalkerbenchmarkingllmsweb}, and WebAggregatorQA.

\vspace{1mm}

\noindent \textbf{Training Configs} \quad
We formalize the trajectory we sampled as $(\text{question}, a_1, o_1, ..., a_n, o_n, \text{answer})$. $a_i$ stands for the \textit{action} code the agent generated to perform tool calling, and $o_i$ is the \textit{observation} returned by the web environment. 
The question and observations are masked during training.

\vspace{1mm}

\noindent \textbf{Baselines and Metrics}\quad
We mainly compare WebAggregator with three types of prior works. \textbf{a.} Non-agentic foundation models that answer questions using their internal knowledge. \textbf{b.} Zero-shot foundation models initialized as agents via the SmolAgents framework. \textbf{c.} Strong fine-tuned foundation models: WebThinker~\citep{wenthinker}, WebDancer~\citep{wu_webdancer_2025}, CognitiveKernel-Pro~\citep{fang2025cognitivekernelpro}, WebSailor~\citep{li_websailor_2025} and WebShaper~\citep{tao2025webshaper}.
We report the pass@1 and pass@3 scores evaluated by GPT-4.1 following previous works~\citep{wu_webdancer_2025}.

\begin{table}[t]
\small
    \centering
    \begin{tabular}{l c | c}
    \toprule
         {\textbf{Model}}& {\textbf{WebWalkerQA}} &{\textbf{XBench}} \\
         \midrule
         {GPT-4.1}  & 39.9  & 37.0  \\
         \hdashline
         
         \multicolumn{3}{l}{\textit{\textbf{WebDancer}}} \\
         {Qwen2.5-7B}  & 36.0  & -  \\
         {Qwen2.5-32B}  & 38.4  & -  \\
         \hdashline
         \multicolumn{3}{l}{\textit{\textbf{WebSailor}}} \\
         {Qwen2.5-7B} & -  &  34.3 \\
         {Qwen2.5-32B} & -  &  53.3 \\
         \hdashline
         
         \multicolumn{3}{l}{\textit{\textbf{WebAggregator}}} \\
         {Qwen2.5-7B}   &{44.7}  & 37.0\\
         {Qwen2.5-32B} & \textbf{45.7}  & {56.0}\\
         {Qwen3-8B} &{41.2}  & {40.0}\\
         {Qwen3-32B} & 43.5  & \textbf{58.0}\\

    \bottomrule
    \end{tabular}
    \caption{Performance on WebWalkerQA and XBench. }
    \label{tab:small-found}
    \vspace{-3mm}
\end{table}

\begin{figure*}[t]
    \centering
    \includegraphics[width=0.9\linewidth]{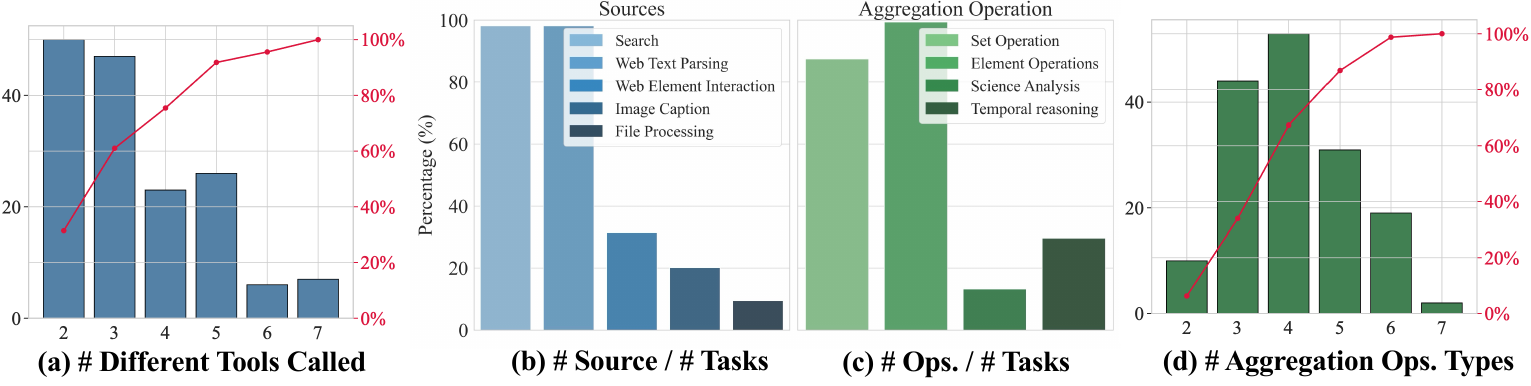}
    \caption{Distributions of tasks required different numbers of tools (a) and aggregation operations (d). Proportion of information source (b) and aggregation operations (c) that are needed across tasks. }
    \label{fig:WebAggregatorQA}
    \vspace{-3mm}
\end{figure*}

\subsection{Main Results}

\textbf{Effects of WebAggregatorQA Training Set} \quad
The experiment results are shown in Table~\ref{tab:related} and Table~\ref{tab:small-found}.
In the zero-shot setting, the closed-sourced models surpass Qwen models on benchmarks.
However, after tuning on WebAggregatorQA, Qwen models exhibit clear and steady improvements on all of the benchmarks and approach the performance of these strong baselines.
Specifically, the WebAggregator based on Qwen2.5-32B and Qwen3-32B surpasses most of the strong baselines, including GPT-4.1 and WebShaper.
The pass@3 performance of WebAggregator-32B achieves 69.9\% on GAIA-text.
These observations prove the quality of WebAggregatorQA and the effectiveness of our data construction paradigm.

\noindent \textbf{Difficulty of WebAggregatorQA Test Set} \quad
The complex aggregation needs in WebAggregatorQA pose a new challenge for current agent systems.
GPT-4.1-powered SmolAgents attain 43.7\% accuracy on GAIA-text but drop to 25.8\% on WebAggregatorQA. 
Claude-3.7-sonnet shows a similar decline. 
Furthermore, the performance gap between Claude and GPT-4.1 is smaller on WebAggregatorQA than on GAIA-text. 
This suggests that for the harder questions in WebAggregatorQA, neither model can solve them effectively, which leads to the reduced gap.
These results and our following analysis (Section~\ref{sec:analysis}) highlight \textbf{the substantial gap between current agent capabilities and the demands of compositional reasoning needed} for multi-hop web tasks.

\noindent \textbf{Data Efficiency and Small Foundations} \quad
The construction of WebAggregatorQA also enables highly efficient model training. By fine-tuning Qwen3-8B on small subsets of 500 and 1,200 trajectories, we achieve 36.9\% and 38.83\% accuracy on GAIA-text, respectively, demonstrating that even limited high-quality data from WebAggregatorQA can yield significant performance gains. This efficiency translates to superior performance in smaller foundation models; for instance, WebAggregator-8B achieves parity with GPT-4.1 on GAIA-text and significantly outperforms previous baselines on WebWalkerQA and XBench (Table~\ref{tab:small-found}). Despite these gains, a performance gap persists between 8B and 32B models on the more complex WebAggregatorQA tasks, suggesting that while small foundations are increasingly viable and cost-effective, breaking through the performance bottleneck of small foundations on hard tasks remains a vital direction.

\vspace{-1mm}
\section{Analysis} \label{sec:analysis}
\vspace{-1mm}

\subsection{Requirements for Solving WebAggregatorQA}
To better understand the challenges in WebAggregatorQA, Figure~\ref{fig:WebAggregatorQA} illustrates the distribution of information sources and the compositional operations required for task completion.

\noindent \textbf{Diversified Information Source Reliance}\quad
Our analysis reveals that tasks in WebAggregatorQA necessitate access to at least Search and Web Text. Furthermore, these tasks frequently require information from Files, indicating a multi-source dependency rather than reliance on a single medium. 
Successfully addressing WebAggregatorQA demands advanced web-browsing capabilities to retrieve knowledge.
The benchmark proves exceptionally challenging for models constrained by internal knowledge alone. Even strong base models like Claude-3.7 and GPT-4.1 solve fewer than 7\% of the questions. 

\noindent \textbf{Diversified Composition Requirements} \quad
We observe that the compositional reasoning requirements of WebAggregatorQA challenge the agent systems.
Figure~\ref{fig:WebAggregatorQA} illustrates that the solving of these tasks requires diversified composition operations, and many of them contain multiple operations.
We conduct further detailed analysis in Section~\ref{sec:core-behavior}.

\subsection{Challenges from Compositional Reasoning}\label{sec:core-behavior}

In this section, we scrutinize the behaviors of research agents and highlight a compound challenge: compositional reasoning on the retrieved information to derive the final answer.

\vspace{1mm}

\noindent \textbf{Success in Seeking, Fail in Reasoning} \quad
We further investigate the impact of compositional reasoning by analyzing a specific failure mode: \textbf{perfect reference retrieval but fails the task}. 
In Table~\ref{tab:agg-ana}, although accuracy improves when all reference URLs are visited, the performance gain remains modest. This suggests that while successful information retrieval is a prerequisite, the subsequent reasoning burden remains a significant bottleneck. 
Notably, Qwen3-32B shows a sharp gap between its strong retrieval and weak accuracy, highlighting a deficit in compositional reasoning. 
WebAggregator-32B, however, significantly closes this gap, reaching performance levels comparable to top-tier baselines. Still, the lack of a high score across all models, even with full information, underscores that the primary ceiling remains the complex composition logic.

\begin{figure}[t]
    \centering
    \includegraphics[width=0.8\linewidth]{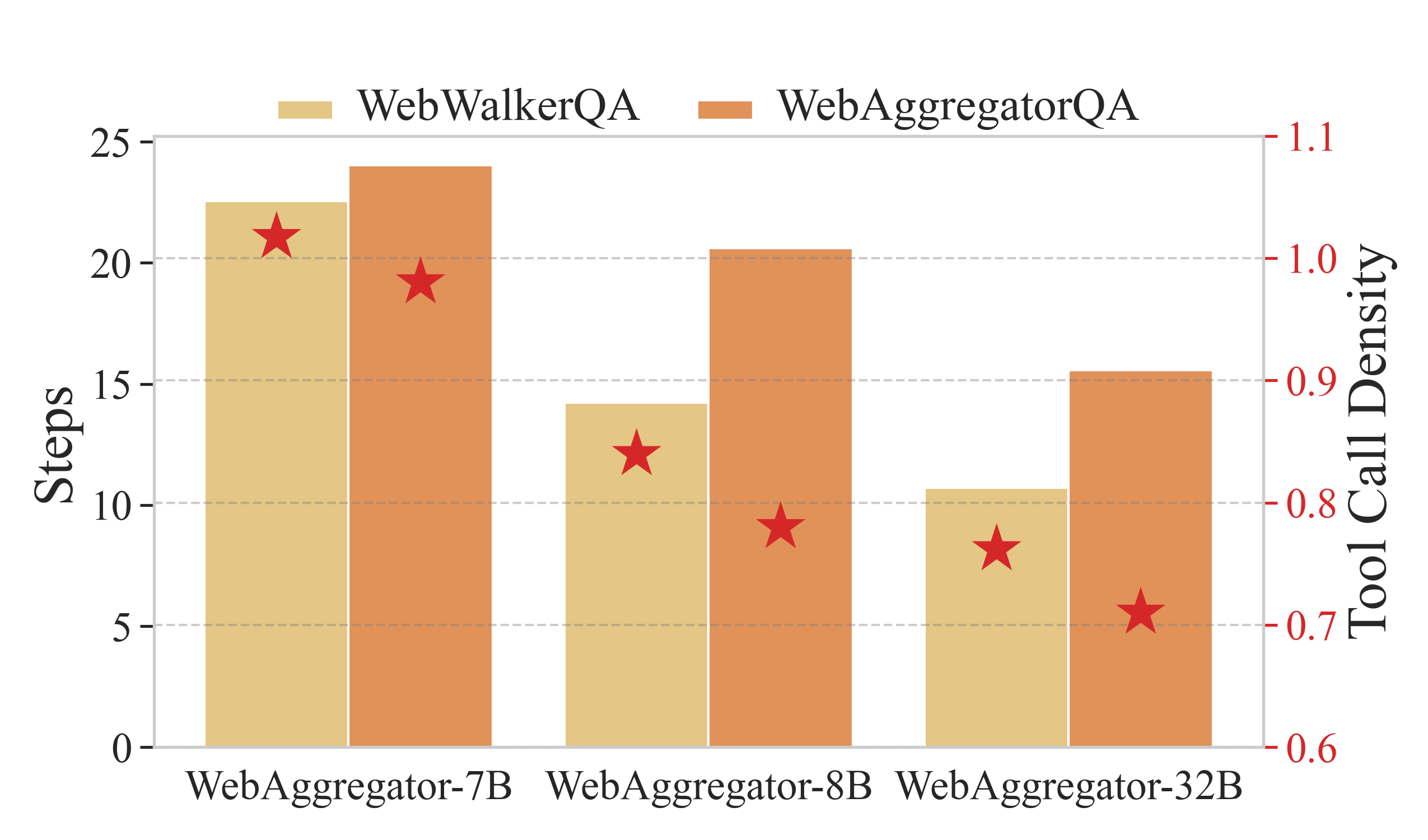}
    \caption{Steps and tool use density of two models across test sets. More results are shown in Table~\ref{tab:model-behav}.}
    \label{fig:tool-use}
    \vspace{-5mm}
\end{figure}

\noindent  \textbf{Reasoning-Intensive Nature of Composition}  \quad
We evaluate the impact of compositional reasoning on agents by analyzing their tool usage patterns across WebAggregatorQA and WebWalkerQA in Figure~\ref{fig:tool-use}. 
We define tool call density as the number of tool calls per step. 
We observe that while tasks in WebAggregatorQA require more total steps to complete, the tool call density is notably lower.
This pattern suggests that in WebAggregatorQA, models rely more heavily on reasoning steps to execute compositional reasoning, enabling deeper synthesis and analysis of existing information, rather than predominantly invoking tools to acquire new external knowledge.

\begin{table}[!t]
\small
    \centering
    \begin{tabular}{lc c}
    \toprule
         {\textbf{Model}} & {\textbf{Counts}} &{\textbf{Acc.}} \\
         \midrule
         \texttt{GPT-4.1} & 30  &  33.3 \\
         \texttt{Claude} & 38  &  42.1 \\
         {\texttt{Qwen3-32B}} &{22}  & {9.7}\\
         \texttt{WebAggregator-32B}  &28  & {35.7}\\
    \bottomrule
    \end{tabular}
    \caption{Counts and accuracy of trajectories that visited all reference URLs and obtained the gold knowledge. }
    \label{tab:agg-ana}
    \vspace{-4mm}
\end{table}

\subsection{Behavior Evolution of WebAggregator}

\noindent \textbf{Evolution of Tool Usage Behaviors} \quad
To understand how fine-tuning on WebAggregatorQA alters agent behavior, we analyzed their action sequences and performed a semantic analysis on $500$ randomly sampled steps per model. 
Following the classification by GPT-5, we categorize internal Thought processes into: 
(1) \textbf{Procedural \& Planning (Type A)}: routine instructions (e.g., "I will search for X") and state updates; 
(2) \textbf{Compositional Reasoning (Type B)}: deep operations including information synthesis, logical deduction, and error correction. 
Our findings, illustrated in Table~\ref{tab:model-behav} and Figure~\ref{fig:too-call-change}, reveal three key trends:

First, \textbf{WebAggregators exhibit reduced tool dependency alongside enhanced reasoning density}. 
As shown in Table~\ref{tab:model-behav}, the Reasoning Ratio (Type B) increased across all scales (e.g., from 18.6\% to 26.2\% for the 8B model), while Tool Call Density markedly declined. This inverse correlation refutes the verbosity hypothesis; rather than generating procedural fluff, the model prioritizes internal logical synthesis over mechanical tool invocation.

Second, \textbf{model scaling further amplifies evidence aggregation capacity}. 
Post-tuning, despite fewer tool calls, total token consumption rose by 12--17\%, a growth driven exclusively by Type B reasoning. This indicates that larger models (32B) leverage deeper internal synthesis, integrating fragmented data into coherent insights, to resolve complex queries with higher efficiency and fewer environment interactions.

Finally, \textbf{agent behavior shifts toward a more diverse and balanced operational distribution}. 
Prior to tuning, tool-calling patterns were concentrated and simplistic (e.g., 40\% of trajectories relied on only two tools, as seen in Figure~\ref{fig:too-call-change}). WebAggregatorQA facilitates a transition from a retrieval-centric paradigm to a reasoning-heavy one, prioritizing multi-step aggregation of scattered evidence.

\begin{figure}
    \centering
    \includegraphics[width=0.8\linewidth]{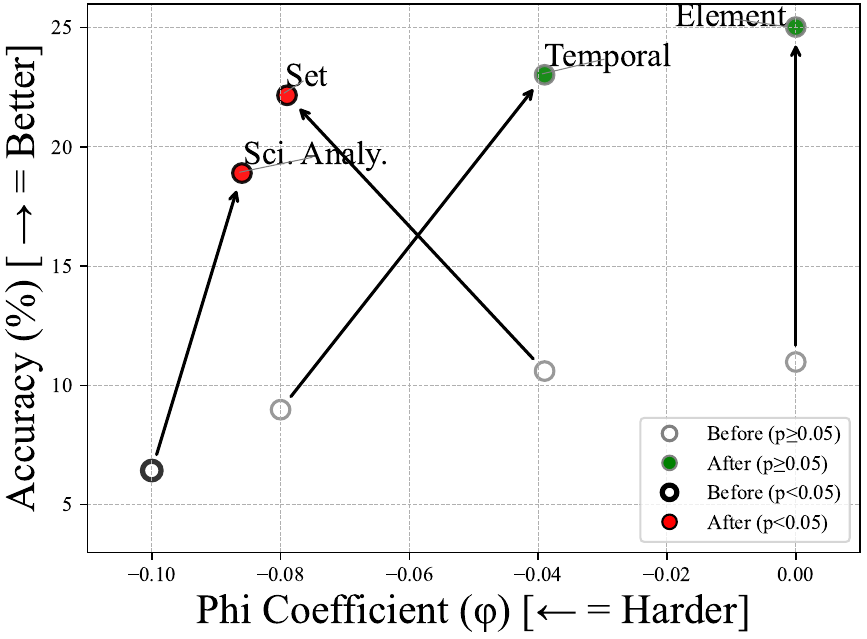}
    \caption{Phi Coefficients ($\psi$) by composition types across models before and after tuning on WebAggregatorQA. }
    \label{fig:logic-change}
    \vspace{-4mm}
\end{figure}

\noindent \textbf{Evolution on Composition Reasoning Capabilities} \quad
{We investigate how model performance changes across different compositional logic types.
Logic labels are labeled by an LLM via analyzing the reference solutions, consistent with the dataset construction method (Figure~\ref{fig:basic-illus}). We compare Qwen series models before and after fine-tuning.}
{We compute the Phi Coefficient ($\psi$), measuring correlation between logic type presence and answer correctness. More negative $\psi$ indicates greater task difficulty. 
Figure~\ref{fig:logic-change} reveals that computation-intensive logic, particularly Scientific Analysis and Set Operations, remains the key challenge.}

\subsection{Screening of Failure Modes}

To elucidate the inherent challenges of WebAggregatorQA, we conduct a qualitative analysis of failure trajectories. We manually examine 50 representative cases from the models listed in Table~\ref{tab:agg-ana} where all references were retrieved, comprising: 28 instances from WebAggregator-32B, 6 each from Claude and GPT-4.1, and 10 additional failure cases. We reveal a taxonomy of three failure categories:

\noindent \textbf{Information Seeking Failures} These errors preclude access to valid evidence. A primary issue is \textbf{Source Location Failure}, where agents misled by initial uncertainty or sub-optimal search queries, fail to locate gold references and settle for incomplete secondary sources.

\noindent \textbf{Agent Foundation Capability Failures}
 Operational errors also derail tasks. Common issues include \textit{Invalid Tool Calls} (e.g., malformed queries) and \textit{Exceeding Max Context}, often resulting from inefficient planning or error-correction loops.

\noindent \textbf{Compositional Reasoning Failures}
{This category covers errors during the composition of retrieved information. 
(1) \textbf{Faulty Composition}: agents misinterpret evidence or apply flawed reasoning, such as making unsupported assumptions, despite accessing the correct sources (Figure~\ref{fig:key-extract}).
(2) \textbf{Error Accumulation}: minor inaccuracies (e.g., premature rounding) compound across steps to yield a significantly incorrect result (Figure~\ref{fig:pre-trunk}). 
(3) \textbf{Ineffective Refinement}: a critical failure where agents fail to self-correct, proceeding down the erroneous logical path while being unable to revise previous mistakes (Figure~\ref{fig:source-finding}).

The above failures are rarely isolated. As shown in Figure~\ref{fig:source-finding}, an initial \textit{Faulty Composition} can propagate errors downstream, exacerbated by \textit{Ineffective Refinement}. This cascade consumes the context window (\textit{Exceeding Max Steps}), increasing the probability of hallucinations or timeouts.

\section{Related Work}

\noindent \textbf{Resources for Deep Research Agents} \quad
Robust information retrieval and precise question answering form the bedrock of deep research agents. Consequently, developing multi-hop datasets for training agent foundation models~\citep{tongyidr,qiao2025webresearcherunleashingunboundedreasoning} necessitates sophisticated tool-use and reasoning, making manual construction challenging. Existing datasets like HotpotQA~\citep{yang-etal-2018-hotpotqa, trivedi2021musique} lack the complexity of authentic web interactions. While some works synthesize request-action pairs~\citep{xu2025agenttrekagenttrajectorysynthesis,chen-etal-2024-agent-flan,he-etal-2024-webvoyager}, they often lack goal-oriented depth. Recent methods construct tasks from logical flows over knowledge snippets~\citep{wu_webdancer_2025,li_websailor_2025,shi_taskcraft_2025,xia2025opendatasynthesisdeep}, such as using KGs from \textit{offline} pages~\citep{shi_taskcraft_2025,wu_webdancer_2025} or entity expansion~\citep{xia2025opendatasynthesisdeep,tao2025webshaper}. However, these remain limited by static content and often neglect knowledge composition (Figure~\ref{fig:basic-illus}), where complexity stems more from reasoning than entity tracing.

\noindent \textbf{Benchmarking Deep Research Agents} \quad
Most existing benchmarks prioritize information-seeking and reference filtering~\citep{wu2025webwalkerbenchmarkingllmsweb,browsercomp-en}, while few~\citep{li2025webweaverstructuringwebscaleevidence} assess compositional reasoning. \citet{frames} evaluates aggregation but is confined to Wikipedia, and \citet{widesearch} focuses on long sequences of simple actions. GAIA~\citep{gaia}, though most relevant, is becoming saturated by recent agents~\citep{fang2025cognitivekernelpro,qiu2025alitageneralistagentenabling}. This underscores the need for challenging benchmarks that jointly evaluate information-seeking and complex reasoning.

\paragraph{Training and Optimization for Research Agents} Recent advancements focus on sophisticated training and test-time optimization. Besides standard rejection-sampling based SFT, which is adopted in most agent post-training works~\citep{tongyidr, fang2025cognitivekernelpro, team2025mirothinker}, there are more advanced techniques leveraging reinforcement learning and test-time self-evolving. In addition to simply applying the training QA pairs as RL input queries, self-synthesized queries can be useful for providing process-level supervision signals for better credit assignment~\citep{fang2025cognitivekernelpro, feng2025gigpo, wang2026spposequencelevelppolonghorizon, li2026verified, fang2026proximity, fu2026maspo}. More specifically, execution errors during tool interaction can also be converted into corrective supervision signals within the RL loop to improve multi-turn tool robustness~\citep{zhang2026robust}. Regarding test-time self-evolving~\cite{fang2025webevolver}, synthesized data could help building agent failure taxonomy and corresponding rubrics for better evaluating the generation of actions~\citep{wan2026inference}.

\section{Conclusion}
In this work, we introduce WebAggregator, a framework designed to bridge the gap between information seeking and compositional reasoning in Deep Research agents. Through the construction of WebAggregatorQA and the WebAggregator models, we demonstrate that verifiable, reasoning-intensive training data significantly boosts agent performance across diverse benchmarks. Crucially, our analysis reveals a fundamental bottleneck: \textit{accurate retrieval does not guarantee task success}, as current models struggle to integrate distributed information. By mitigating these deficits through targeted fine-tuning, WebAggregator paves the way for future research to transcend retrieval-centric paradigms and move closer to autonomous scientific discovery.

\section*{Limitations}
To ensure a fair and controlled comparison, all evaluation protocols and experimental details are strictly aligned with established methodologies~\citep{wu_webdancer_2025,wenthinker,wu2025webwalkerbenchmarkingllmsweb,tao2025webshaper}, precluding any interference from prior knowledge. Following these standards, we evaluated WebAggregator across several well-recognized benchmarks~\cite{gaia,wu2025webwalkerbenchmarkingllmsweb,xbench} to demonstrate its generalization capabilities. Furthermore, we conducted a rigorous data contamination analysis on WebAggregatorQA; as detailed in Appendix~\ref{app:data-contain}, the results confirm zero overlap with existing benchmarks. Given our automated pipeline, we implemented strict post-construction filtering and utilized a rejection sampling procedure during training to eliminate misaligned QA pairs, thereby guaranteeing high data quality. While the data construction of  WebAggregator requires environmental interaction and execution time, it remains highly cost-effective compared to human annotation, which often demands several hours of expert labor per task.
The code, data and models will be open-sourced.
\bibliography{custom,ref}

\clearpage

\appendix
\begin{table*}[!t]
\small
\centering
\begin{tabular}{p{2.8cm} l}
\toprule
\midrule
\textbf{Operations} & \textbf{Questions} \\ 
\hline
\textcolor{blue}{\textit{Element Operations }} & \textit{Aggregate elements/entities, e.g., numbers, times, names$(x, y)$.}  \\
\hdashline
Retrieve $(x)$
& In Amor: A Recipe for Building Adaptable ... , \textcolor{cyan}{what hourly pay (in USD) is} \\
& for the hired NLP expert?  \\
\hdashline

Inverse $(x)$ 
&  \textcolor{cyan}{Which American actor won the Academy Award for XXX in the 1990s} released \\
& their first solo studio album the greatest number of years after their Oscar win? \\
\hdashline

Math $(x,y)$ 
& Among Benedict, Robert Downey, and ..., for the person's first appearance in a\\
&  Marvel Cinematic Universe film corresponded to \textcolor{cyan}{the highest ROI} for their\\ 
&  debut Marvel movie, what is the ROI (three decimals)?\\ 

\hline
\textcolor{blue}{\textit{Set Operations}} & \textit{Aggregate elements $(x)$ and sets $( Y, Z,...)$.}  \\
\hdashline


Filter$(Y)$ & Among the countries that won at least 15 gold medals at the London 2012 Summer\\
&  Olympics, what is the HDI of the country that had \textcolor{cyan}{the third highest} per-capita\\
&  GDP (in USD) in 2012?  \\
\hdashline

Existence$(x,Y)$ &  {For the college that had the most players selected overall in the 2023 NBA Draft,} \\
&  {\textcolor{cyan}{how many} of its draftees were picked in the first round?} \\
\hdashline

Compose$(Y,Z)$ & According to the WorldPopulationReview, how many cities among \textcolor{cyan}{the top 100 most}\\
&  \textcolor{cyan}{populous cities in 2025} have experienced \textcolor{cyan}{a population decrease compared to 2024}? \\

\hline

\textcolor{blue}{\textit{Temporal Reasoning}} & \textit{Reasoning or calculation related with time.} \\
\hdashline
Change 
& Between 1990 and 2022, which country had the third largest average \textcolor{cyan}{annual}   \\
& \textcolor{cyan}{percentage increase} in nominal GDP?\\
\hdashline

TempCalc
&Among Robert De Niro, Al Pacino, Christopher Walken, and Jessica Lange, who    \\
&has the \textcolor{cyan}{longest interval} between their first and most recent Academy Award   \\
& nominations without winning, and \textcolor{cyan}{ what is the length of that span in years?}\\ 
\hline

\textcolor{blue}{\textit{Science Analysis}}  & \textit{Coding is a must to improve efficiency or precision for computation-intensive tasks.} \\
\hdashline
CompIntensive(X) & What is \textcolor{cyan}{the average closure price of Apple.inc from Jan.
2024 to Oct. 2024}?\\
\hdashline
Predict $(x_1, ..., x_n)$ 
&  KFF published an article on abortion in Women's Health Policy on Feb 27, 2025. \\
& Using single exponential smoothing and MSE, search for the optimal alpha  \\
& (0.01-0.99, step=0.01) based on the historical data, the MSE loss, and \textcolor{cyan}{use the alpha to } \\
& \textcolor{cyan}{estimate the next data point}.\\
\hdashline

Statistic $(x_1, ..., x_n)$ &  Among all Cleveland Cavaliers head coaches who have won at least one playoff\\
& game with the team, what is \textcolor{cyan}{the standard deviation} of their playoff win percentages?\\
\hdashline

Correlate $(X,Y)$ 
& Between the 2012 to 2022 NBA seasons, what is \textcolor{cyan}{the Pearson correlation coefficient} \\
& between Damian Lillard's season average points per game and the Portland Trail \\
& Blazers' regular season win percentage?  \\

\midrule
\bottomrule
\end{tabular}
\caption{Several representative examples in WebAggregatorQA of information aggregation operations are presented. Note that the operations here are \textit{high-level guidance} that could be derived into a diversified, specific form, rather than low-level constraints. $x$ means an element or knowledge snippet, $X$ denotes a list of knowledge snippets that fulfill a certain condition.
}
\label{tab:categories}
\end{table*}

\begin{figure*}[htbp]
    \centering
    \includegraphics[width=1\linewidth]{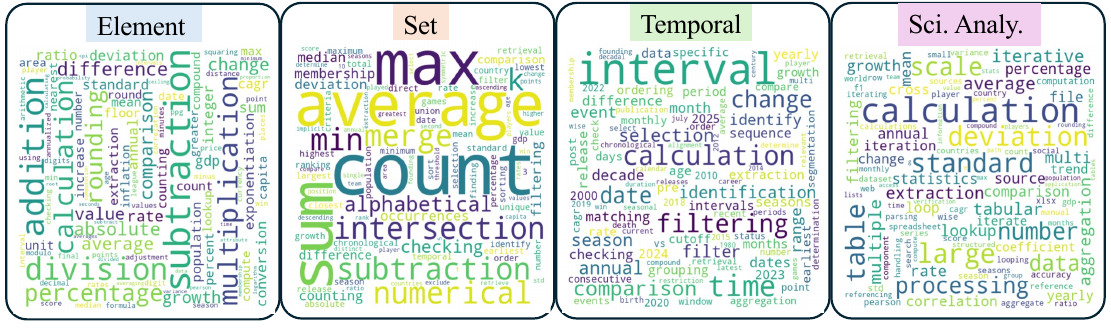}
    \caption{Word cloud of compositional reasoning types extracted from the constructed tasks. High-level logic guidelines are derived into concrete logics.}
    \label{fig:wc}
\end{figure*}

\begin{figure*}[htbp]
    \centering
    \includegraphics[width=0.9\linewidth]{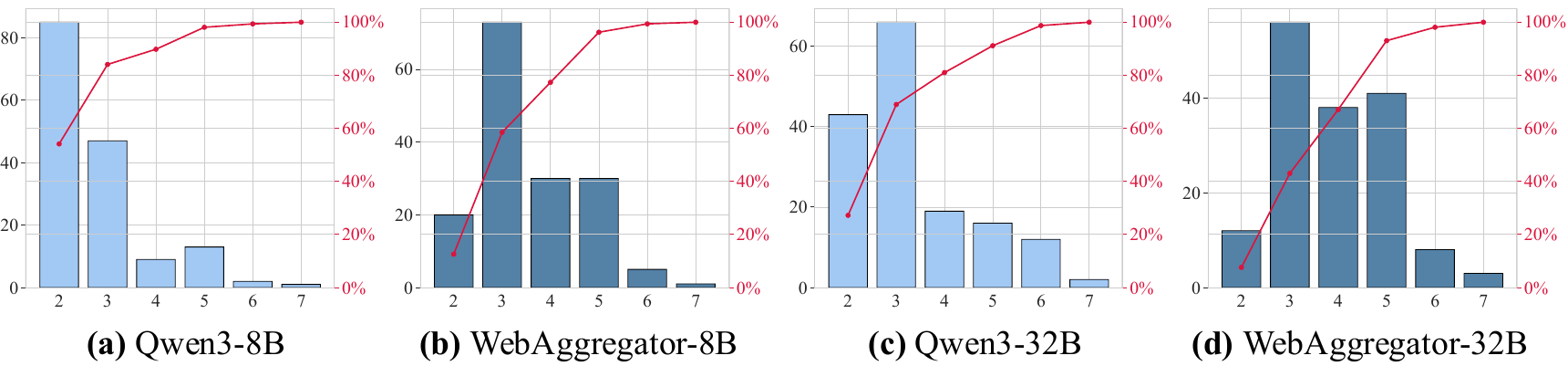}
    \caption{The distribution of unique tools used per trajectory, comparing the model before and after fine-tuning. After fine-tuning, WebAggregator utilizes a more diverse and balanced set of tools within each trajectory. In trajectories that involve only two tools, these are consistently \texttt{Search} and \texttt{Visit}.}
    \label{fig:too-call-change}
\end{figure*}

\begin{table*}[htbp]
\small
    \centering
    \setlength{\tabcolsep}{4pt} 
    \begin{tabular}{l|c|c|c|c|c}
    \toprule
        \textbf{Model} & \textbf{\% TC Steps} & \textbf{TC Density} & \textbf{Total TC} & \textbf{Latency (s)} & \textbf{Reasoning \%} \\
        & (Step-wise) & (TC/Step) & (Sum) & (Avg.) & (Type B) \\
        \midrule
         Qwen3-8B & 90.27\% & 1.52 & 2004 & 301.1 & 18.6\% \\
         \textbf{+WebAggregatorQA(Ours)} & \textbf{86.37\%} & \textbf{0.78} & \textbf{928} & 372.7 & \textbf{26.2\%} \\
         \hdashline
        Qwen3-32B & 91.55\% & 1.12 & 1337 & 225.4 & 31.4\% \\
        \textbf{+WebAggregatorQA(Ours)} & \textbf{82.14\%} & \textbf{0.71} & \textbf{1248} & 306.6 & \textbf{32.6\%} \\
    \bottomrule
    \end{tabular}
    \caption{Behavioral analysis on the WebAggregatorQA test set. WebAggregatorQA training leads to a significant reduction in Tool Call (TC) density and total volume, while increasing the proportion of \textbf{Type B (Reasoning)} steps. This suggests the models shift from mechanical retrieval-centric patterns to more deliberate, compositional deliberation.}
    \label{tab:model-behav}
\end{table*}


\section{Agent Structure}\label{app:agent-st}
\begin{table}
\small
    \centering
    \begin{tabular}{l|l}
    \toprule
         Action & Observation   \\
         \midrule
         \texttt{Search(Query)} & Search results  \\
         
         \texttt{Visit(URL)} & Web Text \& DOM  \\
         \texttt{StrFind(Query)} & Matched str in text\\
         
         \texttt{Input(str, tbox\_id)} & Web Text \& DOM  \\
         \texttt{Click(button\_id)} & Web Text \& DOM\\
         \texttt{Scroll(Pixels)} & Web Text \& DOM\\
         \texttt{Goback()} & Web Text \& DOM\\
         \texttt{FileRead(Path)} & File content\\
         \texttt{Screenshot(Path)} & Capture screen\\
         \texttt{ImageCaption(Path)} & Image description\\
    \bottomrule
    \end{tabular}
    \caption{Action and Observation space.}
    \label{tab:actions-obs}
\end{table}

First, we introduce our agent framework. User instructions often require accessing diverse information on the web—plain text, images, or files—some needing specific interactions like button clicks. 
Thus, the agent must go beyond extracting static page text. We categorize tools by information source: \textbf{Search} (\texttt{Search}), \textbf{Static Page Parsing} (\texttt{Visit}, \texttt{StrFind}), \textbf{Dynamic Interaction} (\texttt{Input}, \texttt{Click}, \texttt{Scroll}, \texttt{Goback}), \textbf{File Processing} (\texttt{FileRead}), and \textbf{Image Captioning} (\texttt{ImageCaption}).

Our implementation utilizes a code-based ReAct~\citep{react} agent, built on SmolAgents~\citep{smolagents}, which outputs natural language thoughts, Python-coded actions, and receives code log as environment observations. 
Each task runs within a fixed 30-step budget, where a step includes an agent action and its observation. We extend the \textit{opendeepresearch} SmolAgents instance with DOM parsing for finer web interactions (see Table~\ref{tab:actions-obs}). This web agent effectively handles tasks requiring information from multiple sources, as summarized in Table~\ref{tab:resource-info}.

\section{More Details for WebAggregatorQA}


\subsection{More Explanations of Quality Control}\label{app:quality-control}
\paragraph{QA Alignment-based Filtering}
To improve data quality, we implemented a two-stage refinement process for task construction. The first stage uses a self-refinement tool with a checklist (Appendix~\ref{app:question-refine-tool}) to quickly verify and revise questions during creation.  Items meeting all criteria are accepted and outputted; those that don't are revised based on feedback until they comply. The second stage involves a data checking agent that thoroughly reviews all reference URLs to ensure alignment of question, answer, and sources (Appendix~\ref{app:prompt-refine-agent}). About 11.72\% of the data were filtered out for failing to meet these standards.

\begin{figure*}
    \centering
    \includegraphics[width=\linewidth]{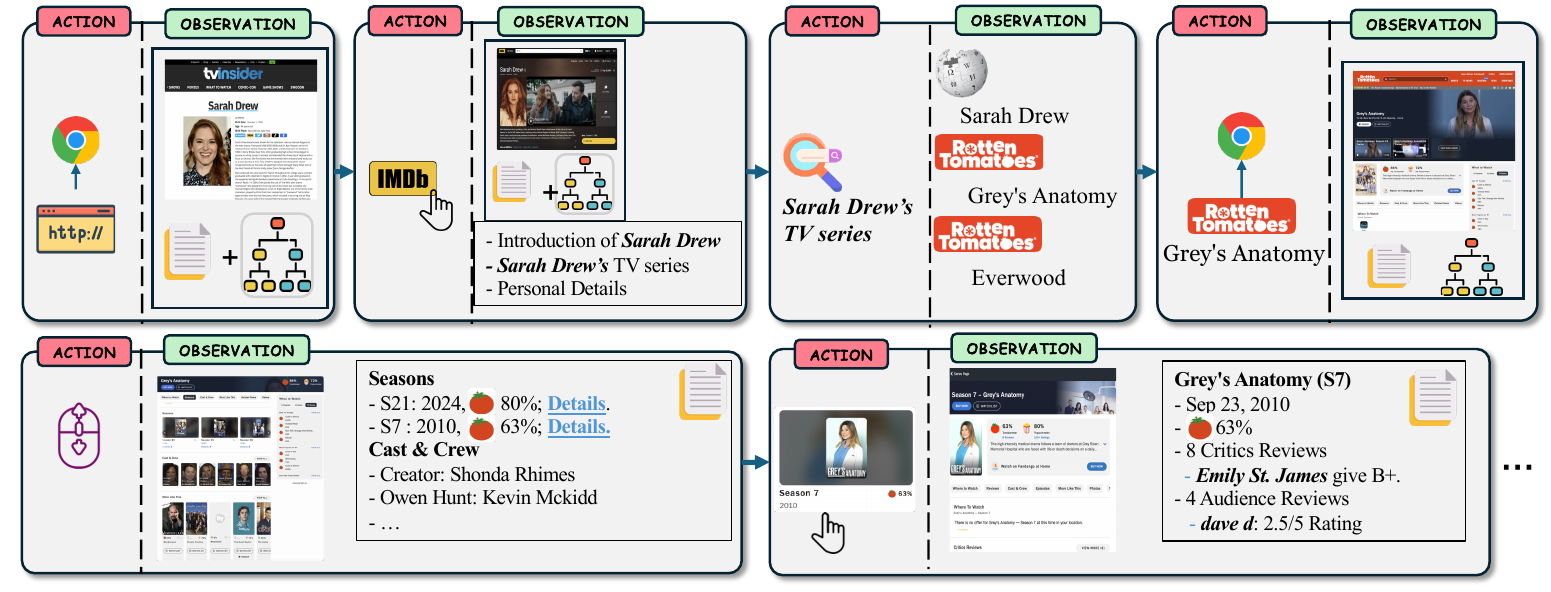}
    \caption{A running example of \textbf{Proactive Web Exploring}: a greater variety of interactions fosters a richer diversity of knowledge and introduces more challenges throughout the process, e.g., questions built from file knowledge also test the file-processing abilities of responding agents.}
    \label{fig:web-explore}
\end{figure*}

\paragraph{Diversity Constraint} 
We ensure dataset diversity by balancing domain and aggregation operation distributions. 
First, we annotated anchor URL domains with GPT-4.1 and sampled data to achieve a more balanced domain distribution, shown in Figure~\ref{fig:dist-domain-step}. 
Second, we analyzed compositional reasoning types using GPT-4.1 to identify high- and low-level operations (e.g., \textit{math subtraction}). 
Although not perfectly reliable without solving the questions, common operations like \textit{calculating average GDP} can be accurately detected. 
We then adjusted prompts to emphasize rare aggregation types, increasing their sample frequency. 
Figure~\ref{fig:wc} shows a word cloud illustrating how different high-level aggregation tasks yield diverse specific operations, such as \textit{intersection} for set operations and \textit{table} processing for Science Analysis.

\paragraph{Data Leakage Avoidance} During proactive web exploration, agents may download and parse existing datasets. 
To prevent data contamination, we created a website keyword blacklist. 
Pages matching the blacklist or containing identified datasets were excluded from retrieval and subsequent model evaluation to ensure the fairness of the evaluation.

\subsection{Testset Annotation} \label{app:test-anno}
\subsubsection{Data Collection and Verification}

The construction of the test set includes the seed tasks collection, several rounds of revision, and cross-validation procedures.
Three human annotators are involved, all of them with at least a bachelor's degree.
Each sample requires an average of 3 hours of annotation work, and the whole procedure lasts for more than 4 weeks of part-time work (4 hours a day).

\textbf{Seed Tasks Collection}\quad
A single human annotator, even if highly specialized in one domain, faces inherent limitations in generating diverse and comprehensive samples across multiple fields. 
To address this, we engage multiple annotators to revise 200 topic-diverse tasks, uniformly selected from WebAggregatorQA based on their domain labels. 
These carefully revised seed examples help ensure that the test set attains the desired diversity.

\textbf{Task Revision Principles}\quad
Our initial analysis, consistent with prior work~\citep{browsercomp-en}, shows that due to high uncertainty in the web environment and an answer-to-question task design, questions are generally well-structured but often lack unique ground truths. 
While this ambiguity may be tolerable for training, it is unacceptable for testing. 
Thus, our key revision principle is to ensure each question is unambiguous and has exactly one correct answer. 

\begin{itemize}[wide=0\parindent,noitemsep, topsep=0.5pt]
\item \textbf{\textit{QA}}
(1) Ensure clarity in the question statements. 
All claims must be explicitly stated, and if multiple sources of evidence exist, additional constraints should be provided in the question to avoid ambiguity. 
(2) The reference answer must be the sole feasible and correct one to the question.
    
\item \textbf{\textit{Reference}} 
Reference information, including URLs and solutions, is vital to the revision process. 
When these reference URLs and solutions are properly validated, the quality and reliability of the questions and answers are assured. 
Accordingly, annotators are required to:
(1) verify the reliability of URLs, ensuring they originate from authoritative and reputable sources;
(2) ensure consistency: the evidence remains stable and not prone to variation across different websites, contexts, or over time;
(3) confirm the fidelity of URLs: each provided reference URL directly and substantively supports the question. Those pages that have a strict CAPTCHA will be replaced with more stable ones. Then the questions and answers are revised accordingly.
    
\end{itemize}

The second principle is to \textbf{increase task complexity} from the same two perspectives: complex compositional reasoning and diversified information sources. 
We provide annotators with the compositional reasoning guidance and encourage them to incorporate more reasoning steps into the questions to enhance their difficulty. 
They are also advised to leverage various information forms beyond plain webpage text.
The answer should not be directly found on the web page.

\begin{figure*}
    \centering
    \includegraphics[width=1.0\linewidth]{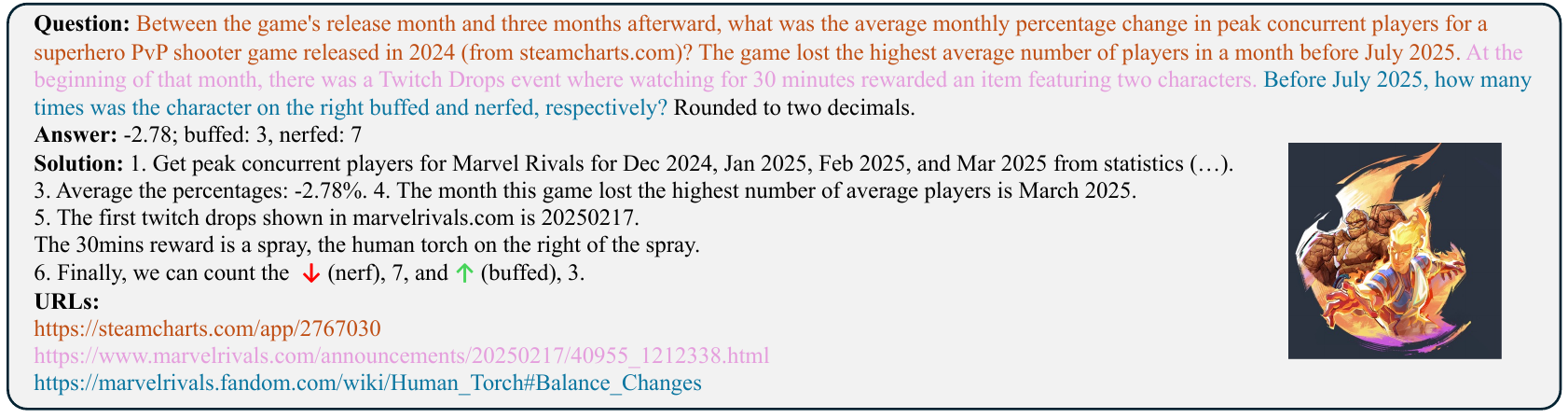}
    \caption{A multimodal sample from the test set of WebAggregatorQA. To solve this task, the agent must extract information from the image to obtain clues for the next step. Since the image is not provided with the question, the agent is required to locate the relevant picture independently.}
    \label{fig:gaia-web-illus-mm}
\end{figure*}

\textbf{Verification}\quad
We utilize agents to assist the human validation procedure. 
Initially, a GPT-4.1-powered agent attempts to solve the questions, facilitating identification of potential ambiguities from the solver’s perspective within a realistic web environment. 
Subsequently, human annotators review the agent’s responses, detect any misunderstandings, and revise the questions, solutions, and answers accordingly. 
This cycle is repeated once more to ensure comprehensive disambiguation.

In the final stage, human annotators independently solve the tasks for cross-validation purposes. Samples that fail to achieve consensus with previously annotated answers are excluded. The independent annotation achieved agreement with 155 out of 159 references, with 4 tasks omitted due to excessive complexity, thereby validating the quality of the references.

\newpage
\onecolumn 

\subsection{Data Construction Prompt} \label{app:data-cons-prompt}

\begin{tcolorbox}[colback=blue!5!white, colframe=blue!75!black, breakable, title = {\textsc{\textit{Part-1}}:  Proactive Explorer}]

\ttfamily
\footnotesize
\textbf{URL:} \\
\{URL\}\\

---

\textbf{Task Overview}

$\bullet$ Create a challenging multi-hop question based on the given URL and related information. \\

$\bullet$ Ensure the quality of the answer when providing a reference answer! Please calculate and verify the reference answer before giving the final data.

$\bullet$ The question should be written in the SAME language as the website content.

---

\textbf{1. Information Gathering}

$\bullet$ Start by thoroughly exploring the given URL and its description.

$\bullet$ Visit and browse at least **\{least\_visits\} different websites** to collect diverse and relevant information.  

$\bullet$ Avoid relying solely on simple search engine queries or Wikipedia. Instead, actively browse, jump between pages, and record your navigation steps and key findings.  

$\bullet$ After each browsing action, briefly document what you did and the important information you discovered.

---

\textbf{2. Question Design }

Formulate a **multi-hop question** that MUST requires reasoning across multiple sources. The answer should **not** be obtainable by a simple search or from a single page. \\

The question should be:

  $\bullet$ Challenging but natural and concise, as if a real user is seeking to learn or solve a puzzle. Avoid unnatural or arbitrary questions such as summing unrelated numbers.
  
  \quad - e.g., year * (number of countries of china) is unacceptable!
  
  $\bullet$ Self-contained.  
  
  \quad - Illustrated with essential clues that guide the respondent to locate the information without explicitly naming the sources or searching queries. The clues must be necessary but precise, avoiding overly broad candidates.
  
  \quad \quad   - BAD EXAMPLES: Some China city has,... (NOT self-contained! Specify the city by specifying the name or providing clues.)
    
  $\bullet$ Based on specific details from at least 5 to 8 different web pages.

  $\bullet$ Reflective of the domain’s characteristics (e.g., medical: functions, gaming: guidance, players, chemistry, math, puzzles).  
  
\end{tcolorbox}

\clearpage

\begin{tcolorbox}[colback=blue!5!white, colframe=blue!75!black, breakable, title = {\textsc{\textit{Part-2}}:  Compositional Logic Proposer}]

\ttfamily
\footnotesize

\textbf{3. Composition Reasoning Operations (Mandatory)  }

Incorporate at least one of the following reasoning operations in your question:

$\bullet$  \textbf{Science Analysis}

> \textbf{Statistical Analysis}  

  \quad - Analyze data from web pages, you may use, but not limited to: calculating the mean, variance, or standard deviation within a specified time period. Some good examples: 
  
    \quad \quad 1. What is the median winnings for drivers who have driven a Chevrolet car?
    
    \quad \quad 2. Which category exhibits the most consistent growth rate across the 5-year period, and what is the average annual percentage increase for that category?
    
    \quad \quad 3. Can you calculate the standard deviation of the average comprehension scores across A, B, and C?

> \textbf{Correlation Analysis}  

  \quad \quad  1. What is the Pearson correlation coefficient (to two decimal places) between China's average annual temperature and its CO2 emissions per capita over the same period?  


> \textbf{Trend Forecasting}  

  \quad - Based on historical data, predict future data points. Any algorithm can be used, such as linear regression, polynomial regression, logistic regression, EMA, etc. REMEMBER: Clearly specify the basis for prediction to ensure a unique answer. Some good examples: 
  
  \quad \quad 1. {Considering the historical data from 1961 to 1967, what could be the forecasted points of Suzuki in the 50cc and 125cc classes for the upcoming years? Use the average growth rate or the most recent 5-year growth rate for prediction.}
  
\quad \quad 2. KFF published an article on abortion in Women's Health Policy on Feb 27, 2025. Using single exponential smoothing and MSE, search for the optimal alpha (0.01-0.99) based on the historical data, the MSE loss, and use the alpha to estimate the next data point.


> \textbf{General Computation Intensive Tasks}  

  \quad - Batch Data Analysis Requires Intensive Computation. The need to retrieve and process large lists of numbers makes coding ESSENTIAL.
  
  \quad \quad 1. What is the average closure price of Apple.inc from Sep. 2024 to Oct. 2024?
  
  \quad \quad 2. Across all NBA seasons where Manu Ginobili's Player Efficiency Rating (PER) exceeded 20 in the regular season, what was the average number of regular season wins by his team?

> \textbf{Other Tasks}  

  \quad - Complex Algorithm with high Complexity: Try to design problems that require coding to reduce time complexity.

------------------------

$\bullet$ \textbf{Element-wise operations}

> Calculation

\quad - Selecting specific elements, performing mathematical operations between elements, e.g., probability, calculation.

  \quad - Examples:  
  
    \quad \quad  1. What is the sum of A’s speed and B’s speed?  
    
    \quad \quad  2. By how much does C’s value exceed D’s value?  
    
    \quad \quad  3. What is the difference between the population of city X and city Y?  

> Inverse Question

\quad - Formalized as an inverse question about certain information. Avoid direct listing; use indirect clues framed as questions. Ensure your phrasing uniquely identifies the subject without ambiguity.

 \quad - Examples:
 
    \quad \quad  1. Instead of "Tom is a singer from New York, who was born on 11 Nov 2024, he...", you can use "for the single from New York, who was born on 11 Nov 2024, he...".
    
    \quad \quad  2. In June 2022, researchers from Huddersfield University published a paper on the application of YOLO in agriculture. My research primarily focuses on ...
    
------------------------

$\bullet$ \textbf{List/Set-wise operations}

\quad > sorting (alphabetical, numerical, top-K), sum, average, counting, intersection, subtraction, merging. Examples:

    \quad \quad 1. Which is the shortest among XXX?  
    
    \quad \quad 2. What is the average length of YYY?  
    
    \quad \quad 3. How many items appear in both set A and set B?  
    
    \quad \quad 4. What is the total number of Z across all categories? 

------------------------

$\bullet$ \textbf{Element-Set operations}

\quad > checking membership or counting occurrences.
Examples:  
  
    \quad \quad 1. Is element E part of the top 10 ranked items?  
    
    \quad \quad 2. Exclude all names that were born in 1984 from ... 
    
    \quad \quad 3. Between 2012 and 2021, was the rate of increase in China's average annual temperature higher or lower than the global average?

    \quad \quad 4. On the same day that a landmark house on South Main Street in Coeymans Landing, New York, rich with local history, built in the late 1830s, officially entered the National Register of Historic Places listing, how many places entered the list total?
    
------------------------

\textbf{Note:} The numbers or elements used in these operations should be discoverable by reading the web content, not directly provided in the question.

------------------------

\textbf{4. Answer Requirements } 

  > The answer MUST not be obtained directly from the retrieved text and MUST be derived through reasoning.
  > Short, Concise and easy to verify.  
  
  > Stable over time (avoid dynamic or real-time data). 
  
  > Of a clear entity type (e.g., person, number, date, place).  

------------------------

\textbf{5. Output Format } 

Output your final result in the following JSON format:

\{

  \quad "topic": "Brief description of the question’s domain or topic",
  
  \quad "question": "The constructed multi-hop question",
  
  \quad "answer": "The answer X",
  
  \quad "context": \{

    \quad "URLs": [
      "url\_1",
      "url\_2",
      "url\_3",
      "url\_4",
      "url\_5",
...
    ]
    
  \quad \}
  
\}

\end{tcolorbox}

\clearpage

\subsection{Prompt of Data Quality Checking Agent} \label{app:prompt-refine-agent}

\begin{tcolorbox}[colback=blue!5!white, colframe=blue!75!black, breakable, title = {\textsc{Task Description of Data Quality Checking Agent}}]

\ttfamily
\footnotesize
\{{Composition Reasoning Operations Prompt}\}

Evidence Checking

\begin{todolist}
\item \textbf{URL Validity}: Verification that all URLs conform to proper syntax and resolve correctly without errors.
\item \textbf{Information Relevance}: Assessment of whether each URL contains information that is necessary and sufficient to address the research question.
\end{todolist}

Question Checking

\begin{todolist}
\item \textbf{Self-Containment}: The extent to which the question is fully specified and comprehensible without requiring additional external context.
\item \textbf{Retrieval Necessity}: The degree to which answering the question necessitates consulting external sources, while avoiding excessive disclosure of information within the question itself.
\item \textbf{Composition Necessity}: The question must include at least three different aggregation operations, ensuring that the answer cannot be obtained through direct retrieval.
\item \textbf{Clarity}: The precision and unambiguity of the cues or references embedded in the question that facilitate accurate data retrieval. The clues will not lead to multiple feasible answers.
\item \textbf{Temporal Stability}: The property that the correct answer to the question remains consistent over time, unaffected by temporal changes (e.g., “Who was the immediate past president of the United States?”).
\end{todolist}

Answer Quality Assessment

\begin{todolist}
    \item \textbf{Information Fidelity}: The extent to which all information presented in the reference answer is fully consistent with the URLs or other provided external information sources.  
    \begin{itemize}
        \item \textit{Example of inconsistency}: The temperature retrieved from the reference URL is 37°C, whereas the solution states 35°C, resulting in an erroneous calculation of the average temperature.
    \end{itemize}
    
    \item \textbf{Ground Truth Validity}: The reference answer must accurately and unambiguously reflect the requirements of the question, conforming to information obtained from authoritative and reliable data sources.  
    \begin{itemize}
        \item The answer should be derived from recognized authoritative channels or verified databases.
        \item Ensuring verifiability through reliable sources is especially important for questions involving numerical data, statistics, or other factual information.
        \item \textit{Example of invalid answer}: “The moon’s distance from Earth is 100,000 km.” This contradicts scientific consensus, which states the distance is approximately 384,400 km.
    \end{itemize}
    
    \item \textbf{Uniqueness and Unambiguity}: The reference answer should be uniquely correct, avoiding ambiguity or multiple plausible solutions.  
    \begin{itemize}
        \item Are there conflicting data from multiple sources that lead to multiple possible answers?
        \item Are there precision conflicts between different data sources (e.g., 33.2 vs. 33.20987)?
    \end{itemize}
\end{todolist}

----------------------

\textbf{Based on the above criteria, analyze the following data by solving it:}

\textbf{Question}: \{\}

\textbf{Answer}: \{\}

\textbf{Evidence\_URLs}: \{\}

Please verify whether each item meets the standards.

----------------------

\textbf{Output Format}

Return your analysis in the following JSON format:

```json

\{
    
    \quad "Evidence Passed": 1 or 0,

    \quad "Question Passed": 1 or 0,

    \quad "Answer Passed": 1 or 0,
  
  \quad "Domain": "[USE ONLY ONE WORD OF THE FOLLOWING!] Gaming, Sport, TV shows \& movies, Computer Science, Art, History, Music, Geography, Politics, Finance, Medical, Law",
  
  \quad "Compositional\_Operation": 
  
  \quad \{
  
    \quad "type": ["Science Analysis Operations->Informations search->XLSX Processing of ...", "Element-wise->Math->Addition", "Science Analysis Operations->Batch Data Processing->", ...]
    
  \quad \}
  
\}

```
\end{tcolorbox}

\clearpage

\subsection{Prompt of Intergrated Data Quality Checking Tool} \label{app:question-refine-tool}

\begin{tcolorbox}[colback=blue!5!white, colframe=blue!75!black, breakable, title = {\textsc{Prompt of Efficient Question Checking Tool}}]

\ttfamily
\footnotesize
\{{Composition Reasoning Operations Prompt}\}

Question Checking

\begin{todolist}
\item \textbf{Self-Containment}: The extent to which the question is fully specified and comprehensible without requiring additional external context.
\item \textbf{Retrieval Necessity}: The degree to which answering the question necessitates consulting external sources, while avoiding excessive disclosure of information within the question itself.
\item \textbf{Compositional Necessity}: The question must include at least three different aggregation operations, ensuring that the answer cannot be obtained through direct retrieval.
\item \textbf{Clarity}: The precision and unambiguity of the cues or references embedded in the question that facilitate accurate data retrieval. The clues will not lead to multiple feasible answers.
\item \textbf{Temporal Stability}: The property that the correct answer to the question remains consistent over time, unaffected by temporal changes (e.g., “Who was the immediate past president of the United States?”).
\end{todolist}

----------------------

Based on the above criteria, analyze the following data:

\textbf{Question}: \{\}

\textbf{Answer}: \{\}

\textbf{Evidence\_URLs}: \{\}

Please verify whether each item meets the standards and provide advice for improvements.

\end{tcolorbox}

\newpage
\twocolumn 

\begin{figure*}[htbp]
    \centering
    \includegraphics[width=0.95\linewidth]{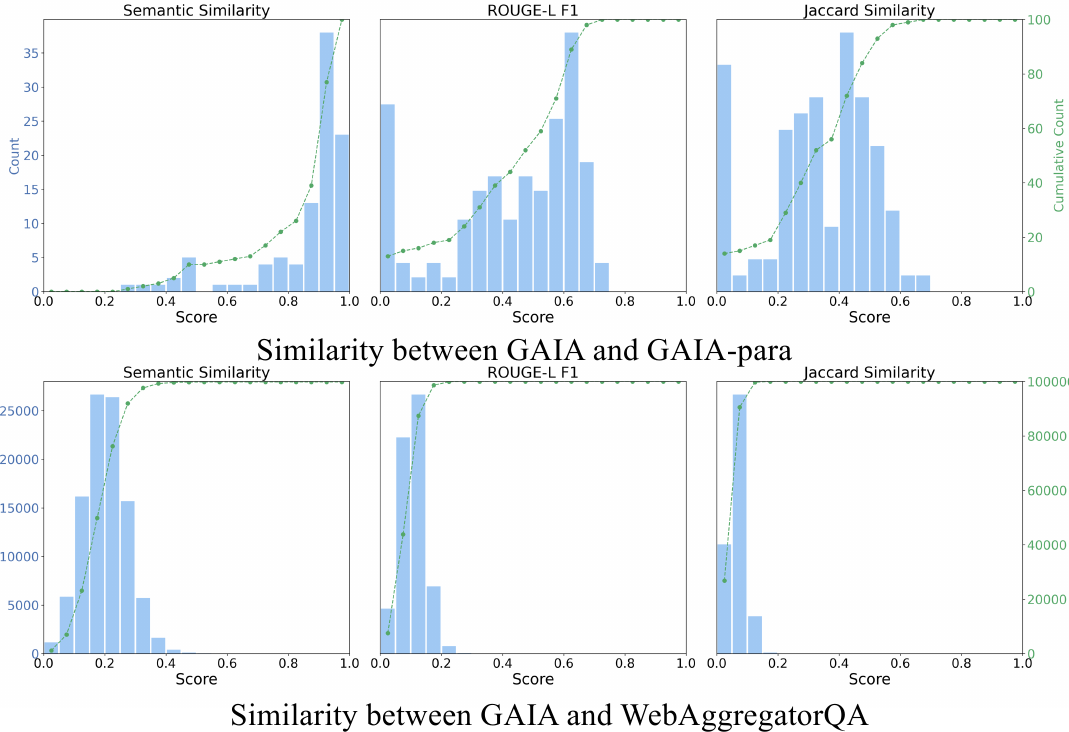}
    \caption{Data Contamination Experiment}
    \label{fig:data=con-analysis}
\end{figure*}
\section{Data Contamination Analysis}\label{app:data-contain}

{We investigate potential data contamination between WebAggregatorQA and GAIA through two similarity analyses, with the results presented in Figure~\ref{fig:data=con-analysis}. First, to establish a baseline, we prompt an LLM to paraphrase the GAIA dataset, creating a new version termed GAIA-para. We then measure the similarity between the original and paraphrased GAIA questions. Second, we assess the cross-dataset overlap by sampling 100,000 pairs from the Cartesian product of the WebAggregatorQA training set and GAIA and computing their similarity. For all comparisons, we employ three metrics: a semantic similarity score calculated by Qwen3-Embedding-8B, ROUGE-L, and Jaccard similarity.}

{Figure~\ref{fig:data=con-analysis} reveals a stark contrast: while GAIA shows significant similarity with its paraphrased counterpart, its overlap with WebAggregatorQA is negligible. This provides strong evidence that \textbf{our WebAggregatorQA dataset is free from GAIA data contamination}.}

\section{Close Screening Agent Trajectories}

\newpage
\onecolumn 
\subsection{Example 1} \label{fig:key-extract}
\begin{chatconversation}[]{\textcolor{red}{Example 1}}\

    \centering

    \begin{userbox}
        Between 2014 and 2024, which year has the most new Trustees of the  current Embry-Riddle Board of Trustees? And how many of the current Trustees received the Living Legend of Aviation?
    \end{userbox}

    \begin{agentbox}
        First, I need to find the official Embry-Riddle Aeronautical University Board of Trustees page. This will likely contain the current list of Trustees and possibly historical data or links to archives. I will search for the Board of Trustees page and review its contents for the necessary information.

\begin{lstlisting}[language=Python, caption={}, label={lst:d-2}]
results = web_search(query="Embry-Riddle Aeronautical University Board of Trustees")

print(results)
\end{lstlisting}

    \end{agentbox}
    
    \begin{userbox}
        1. [Embry-Riddle Aeronautical University: Board of Trustees](https://trustees.erau.edu/)
        
        2. [Meet the Trustees](https://trustees.erau.edu/trustees)
        
        3. [Trustee Ken Dufour and Chairman Mori ...](https://news.erau.edu/headlines/trustee-ken-dufour-and-chairman-mori-hosseini-mark-milestones-on-embry-riddles-board)
        
        4. ...
    \end{userbox}

        \begin{agentbox}
The top search result is the official Embry-Riddle Board of Trustees page: https://trustees.erau.edu/. 

This page is likely to have a current list of Trustees and possibly links to historical data or news about new appointments. 

I will visit this page and review its contents to gather the information needed, especially looking for sections about current Trustees, their appointment years, and any links to archives or news about new appointments.

\begin{lstlisting}[language=Python, caption={}, label={lst:d-3}]
board_page = visit_webpage(url="https://trustees.erau.edu/")

print(board_page)
\end{lstlisting}
    \end{agentbox}
    \begin{userbox}
        ...
    \end{userbox}
\begin{agentbox}
Since there is no unified list or page displaying the year each trustee became a trustee, I will open each trustee's profile one by one to check and confirm whether they have received the Living Legend of Aviation award.
\begin{lstlisting}[language=Python, caption={}, label={lst:d-4}]
# Visit the papes of the first ten trustees
urls=[
"https://trustees.erau.edu/trustees/michael-amalfitano",
"https://trustees.erau.edu/trustees/michael-amalfitano",
"https://trustees.erau.edu/trustees/leanne-caret",
...
]
print(len(urls)) # There are {20} pages
for url in urls[:10]:
    print(visit_webpage(url))
\end{lstlisting}

    \end{agentbox}
    \begin{userbox}
    
    ...
    
Mori Hosseini is the Chairman and CEO of ICI Homes, one of Florida’s largest residential homebuilder/developers. Born in Iran, Mr. Hosseini was educated in London, England, ...

- Mori Hosseini (HonDoc '13; '78, '79, '82, DB)

- Member since \textbf{1999}

- Chair, Board of Trustees since 2014

- Chair, Committee on Trustees

...
    \end{userbox}
\begin{agentbox}
I will continue to access other pages and record the years they became members.
\begin{lstlisting}[language=Python, caption={}, label={lst:d-5}]
trustees=[
"Michael Amalfitano" :{ 
 "member year": "2021",
 "living legend": False
},

"John Amore" :{ 
 "member year": "2010",
 "living legend": False
},
...
"Mori Hosseini" :{ 
 "member year": "2014",  
 "living legend": False
},
...
]
for url in urls[10:]:
    print(visit_webpage(url))
\end{lstlisting}

\begin{expblock}{Error Analysis: Faulty Composition}
     Key information extraction error: \textcolor{red}{2014 v.s. 1999}
\end{expblock}
    \end{agentbox}
\begin{userbox}
    ...
\end{userbox}
\begin{agentbox}

\begin{lstlisting}[language=Python, caption={}, label={lst:d-7}]
final_answer=(\{"year":2014, "number of awardees": 3 \}) 
\end{lstlisting}

\begin{expblock}{Error Analysis: Faulty Composition}
     Faulty Composition while \textit{Set->Filter->Counting}
\end{expblock}

\end{agentbox}

\begin{userbox}
END>> answer: 2014, 3  \textcolor{red}{\# true answer: 2014\&2018, 2}
\end{userbox}

\end{chatconversation}




% ...


    




% ...





    






\subsection{Example 2} \label{fig:source-finding}

\begin{chatconversation}[fig:chat-breakable]{\textcolor{red}{Example 2}}

    \centering

    \begin{userbox}
After one game's update introducing the Embrion and Old Bird, calculate the compound monthly CAGR of its average players from the update's release month until (but excluding) the next major patch's month, using steambase.io data. For the same period, calculate the CAGR of another game, which is a 4 player online co-op psychological horror game released in 2020. Present both game name and number as 'game: ±XY.ZT\%'.
    \end{userbox}

    \begin{agentbox}
Step 1: Identify the game that received an update introducing "Embrion" and "Old Bird". I will use a web search to find which game this is and get the update's release month. After that, I will look for the next major patch's release month for that game.

\begin{lstlisting}[language=Python, caption={}, label={lst:d-8}]
search_query = 'game update introducing Embrion and Old Bird'

results = web_search(query=search_query)

print(results)
\end{lstlisting}

    \end{agentbox}
\begin{userbox}
...
\end{userbox}
    \begin{agentbox}
The update introducing Embrion and Old Bird is Version 50 of Lethal Company, released on April 13, 2024. I now need to find the release month of the next major patch after Version 50. I will search for "Lethal Company next major patch after Version 50" to determine this.

\begin{lstlisting}[language=Python, caption={}, label={lst:d-9}]
next_patch_search = web_search(query="Lethal Company next major patch after Version 50")

print(next_patch_search)
\end{lstlisting}

\end{agentbox}

\begin{userbox}
...
\end{userbox}

\begin{agentbox}
The most relevant sources suggest that after Version 50, \textcolor{red}{the next major patch is speculated to be Version 70}, but there is no explicit mention of an official release date. Since the question specifies to use steambase.io data, I should check steambase.io for the version history or patch notes for Lethal Company to see if there is a clear next major patch after Version 50. If not, I might have to use the next version listed or clarify if there is a clear, next major patch. I will visit the steambase.io Lethal Company page and look for update/version history.

\begin{lstlisting}[language=Python, caption={}, label={lst:d-10}]
lethal_company_steambase = web_search(query="Lethal Company steambase.io")

print(lethal_company_steambase)
\end{lstlisting}

\begin{expblock}{Error Analysis: Faulty Composition}
    The next major patch is v60, on August 17, 2024. But the agent takes a guess.
\end{expblock}

\end{agentbox}

\begin{userbox}
...

\end{userbox}

\begin{agentbox}
The search did not provide a direct link to the Lethal Company page on steambase.io, but it did mention steambase.io as a source for player counts. I should visit steambase.io and look for the Lethal Company page, or try to directly access the relevant URL. Once there, I can look for monthly average player counts for the period starting April 2024 (the release month of Version 50) and \textcolor{red}{check for any next major patch (likely Version 70)} or any other official patch that is listed as "major" after Version 50.

\begin{lstlisting}[language=Python, caption={}, label={lst:d-11}]
lethal_company_url = "https://steambase.io/games/lethal-company"

page_info = visit_webpage(url=lethal_company_url)

print(page_info)
\end{lstlisting}

\begin{expblock}{Error Analysis: Underperformed Iterative Refinement}
    The next major patch is `v60`, on August 17, 2024. But the agent no longer checks whether `v70` is reasonable and proceeds with subsequent steps.
\end{expblock}

\end{agentbox}
\begin{agentbox}
    The update introducing Embrion and Old Bird (Version 50) was released in April 2024. The next major patch after Version 50 is not explicitly listed, but the next major patch after 50 is Version 70, which is referenced in previous search results as the next major update. 
    
    ...
    
    Now, I will extract the average player counts for Lethal Company from April 2024 to March 2025 (12 months, not including April 2025), as that's the period from the update's release month until (but not including) the next major patch's month.
    \begin{lstlisting}[language=Python, caption={}, label={lst:d-12}]
# Monthly average players for Lethal Company from steambase.io (April 2024 - March 2025)
    lethal_company_averages = {    "2024-04": 24020,    "2024-05": 13345,    "2024-06": 12073,    "2024-07": 14299, 
    
    "2024-08": 16523,    "2024-09": 16036,    "2024-10": 13545,    "2024-11": 14299,  
    
    "2024-12": 13120,    "2025-01": 17462,    "2025-02": 15399,    "2025-03": 8886}
    
print(lethal_company_averages)
\end{lstlisting}

    \begin{expblock}{Error Analysis: Underperformed Iterative Refinement}
    The next major patch is `v60`, on August 17, 2024. But the agent no longer checks whether `v70` is reasonable and proceeds with subsequent steps.
\end{expblock}

\end{agentbox}

\begin{userbox}
    \textcolor{red}{The rest of the trajectory is omitted for readability.}
\end{userbox}
\end{chatconversation}

\subsection{Example 3} \label{fig:pre-trunk}
\begin{chatconversation}[]{}
    \begin{userbox}
Using data from the World Bank for 2014-2023, calculate annual averages for: (1) GDP growth rate, (2) Inflation rate (consumer prices), (3) CO2 emissions per capita growth (excluding LULUCF), (4) Internet users (\% population), (5) Women in parliament (\%). Find the intersection between: (a) top 200 countries in (1),(3),(4),(5) and (b) bottom 200 countries in (2). Rank the countries in ascending order based on their CO2 emissions growth rate, with CO2 emissions growth rate rounded to 5 decimals. For the growth rate conditions mentioned above, if the World Bank provides growth rate data, prioritize using the World Bank's growth rate data. Otherwise, calculate it using the following formula: Growth Rate (\%) = ( (data[n] - data[n-1]) / (data[n-1] + 0.00001) ) × 100  The small constant 0.00001 is added to avoid division by zero in cases where data[n-1] = 0. Give me the answer as 'Country:  CO2 emissions growth rate'
    \end{userbox}

    \begin{agentbox}
This is a comprehensive data-rich task involving 5 World Bank indicators for all countries from 2014-2023. I need to:

1. Identify the exact World Bank indicator codes for each metric.

2. Download the datasets for each indicator (preferably from the World Bank API or CSV download links).

3. For each country and indicator, compute the annual averages for the period 2014-2023.

4. For indicators (1), (3), and (2): if WB has growth rate data, use it; if not, compute it using the formula provided.

5. Select the top 200 countries in (1) GDP growth, (3) CO2 per capita growth, (4) Internet users, (5) Women in parliament, and bottom 200 in (2) inflation. Find the intersection countries among top 200 in 1/3/4/5 and bottom 200 in 2.

6. Rank those by ascending order of their CO2 emissions growth rate, rounded to 5 decimals.  

This is a multi-step data task. First, I will search for the official World Bank indicator codes for each variable.

\begin{lstlisting}[language=Python, caption={}, label={lst:data_processing_functions}]
queries = [

"World Bank indicator for GDP growth rate",

"World Bank indicator for inflation rate (consumer prices)",

"World Bank indicator for CO2 emissions per capita growth (excluding LULUCF)",

"World Bank indicator for Internet users percent the population",

"World Bank indicator for women in parliament percent"

]

for q in queries:
    print(f"{q}:", web_search(query=q))
\end{lstlisting}

\end{agentbox}

    


    

\begin{userbox}
    ...
\end{userbox}

\begin{agentbox}
I'll process the data **directly in memory**, reading from the zipfile into a `csv.reader` and printing the header/first lines for inspection.  \\ This approach requires no file system operations and should work given the allowed modules.

\begin{lstlisting}[language=Python, caption={}, label={lst:data_processing_functions-1}]
def average(values):
    nums = [float(v) for v in values if v not in ('', None)]
    return sum(nums)/len(nums) if nums else None

def CO2_growth_rate(vals):
    # compute per-year growth rates, then average them
    result = []
    # Need at least two years to compute one growth rate
    vals_num = [float(v) if v not in ('', None) else None for v in vals]
    for i in range(1, len(vals_num)):
        if vals_num[i] is not None and vals_num[i-1] is not None:
            prev = vals_num[i-1]
            curr = vals_num[i]
            # formula with epsilon to avoid division by zero
            rate = ((curr - prev) / (prev + 0.00001)) * 100
            result.append(rate)
    return sum(result)/len(result) if result else None

indicator_avgs = {}
for key in indicator_data:
    indicator_avgs[key] = {}
    for code, vals in indicator_data[key].items():
        if key == "CO2_pc":
            avg = CO2_growth_rate(vals)
        else:
            avg = average(vals)
        indicator_avgs[key][code] = avg

# Print samples for all indicators to verify
for k in indicator_avgs:
    print(f"Averages for {k}:", list(indicator_avgs[k].items())[:3])
\end{lstlisting}


















\end{agentbox}
\begin{userbox}
    ...
\end{userbox}
\begin{agentbox}
I will obtain the countries with the highest CO2 emissions growth rates. Then I will begin calculating other metrics and finding their intersections.

    \begin{lstlisting}[language=Python, caption={}]
# Prepare (country, rounded growth rate) pairs for intersection countries
results = []
for code in intersection:
    val = indicator_avgs['CO2_pc'][code]
    if val is not None:
        results.append((code, round(val, 5)))

# Sort by CO2 emissions growth rate, ascending
results_sorted = sorted(results, key=lambda x: x[1])

# Obtain the top 200 countries and growth rate
country_rate=results_sorted[:200]
print(country_rate)
\end{lstlisting}

\begin{expblock}{Error Analysis: Error Accumulation in Composition}
    Due to \textbf{the pre-reduction of the country list in this step}, the final intersection differed from the actual results, causing \textbf{this trajectory to ultimately return an empty list}.
\end{expblock}
\end{agentbox}
\end{chatconversation}

\end{document}